\documentclass[10pt,twocolumn,letterpaper]{article}

\usepackage[pagenumbers]{wacv} %

\definecolor{wacvblue}{rgb}{0.21,0.49,0.74}

\makeatletter
\newcommand{\copyrightnotice}[1]{\gdef\@copyrightnotice{#1}}
\newcommand{\toappear}[1]{\gdef\@toappear{#1}}
\let\@copyrightnotice\@empty
\let\@toappear\@empty
\newcommand{\ps@myfooter}{%
  \let\@mkboth\@gobbletwo
  \def\@oddhead{}%
  \def\@evenhead{}%
  \def\@oddfoot{\rlap{\@toappear}\hfil\thepage\hfil\llap{\@copyrightnotice}}%
  \let\@evenfoot\@oddfoot
}
\makeatother

\toappear{\it To appear in Proc.\ WACV 2026 Workshops}
\copyrightnotice{\copyright\ 2026 IEEE}
\usepackage{ifthen}
\usepackage{atbegshi}
\newcounter{pagecounter}
\AtBeginShipout{%
  \stepcounter{pagecounter}%
  \ifthenelse{\value{pagecounter}=1}{%
    \AtBeginShipoutUpperLeft{%
      \put(\dimexpr\paperwidth-20.3cm\relax,-0.8cm){%
        \parbox[t]{19cm}{%
          \copyright\ 2026 IEEE. Personal use of this material is permitted. 
          Permission from IEEE must be obtained for all other uses, in any current or future media, including reprinting/republishing this material for advertising or promotional purposes, creating new collective works, for resale or redistribution to servers or lists, or reuse of any copyrighted component of this work in other works.%
        }%
      }%
    }%
  }{}%
}

\usepackage[pagebackref,breaklinks,colorlinks,allcolors=wacvblue]{hyperref}
\usepackage{multirow,multicol}
\usepackage{pifont}

\title{Detector-Augmented SAMURAI for Long-Duration Drone Tracking}

\author{
    Tamara R. Lenhard$^{1,2,3}$ 
    \and
    Andreas Weinmann$^{2}$
    \and
    Hichem Snoussi$^{3,4}$
    \and
    Tobias Koch$^{1}$\\\vspace{-0.4cm}\and
    {\normalsize$^{1}$Institute for the Protection of Terrestrial Infrastructures,
    German Aerospace Center (DLR), Sankt Augustin, Germany}\\
    {\normalsize$^{2}$ACIDA Lab, Technical University of Applied Sciences Würzburg-Schweinfurt, Schweinfurt, Germany}\\
    {\normalsize$^{3}$Data Science Institute, European University of Technology, European Union}\\
    {\normalsize$^{4}$LIST3N, Université de Technologie de Troyes, Troyes, France}\\\vspace{-0.4cm}\and
    {\tt\small \{tamara.lenhard, tobias.koch\}@dlr.de, andreas.weinmann@thws.de, hichem.snoussi@utt.fr} \vspace{-0.4cm}
}

\begin{document}
\maketitle

\begin{abstract}
Robust long-term tracking of drone is a critical requirement for modern surveillance systems, given their increasing threat potential. While detector-based approaches typically achieve strong frame-level accuracy, they often suffer from temporal inconsistencies caused by frequent detection dropouts. Despite its practical relevance, research on RGB-based drone tracking is still limited and largely reliant on conventional motion models. Meanwhile, foundation models like SAMURAI have established their effectiveness across other domains, exhibiting strong category-agnostic tracking performance. However, their applicability in drone-specific scenarios has not been investigated yet. Motivated by this gap, we present the first systematic evaluation of SAMURAI's potential for robust drone tracking in urban surveillance settings. Furthermore, we introduce a detector-augmented extension of SAMURAI to mitigate sensitivity to bounding-box initialization and sequence length. Our findings demonstrate that the proposed extension significantly improves robustness in complex urban environments, with pronounced benefits in long-duration sequences -- especially under drone exit--re-entry events. The incorporation of detector cues yields consistent gains over SAMURAI’s zero-shot performance across datasets and metrics, with success rate improvements of up to $+$0.393 and FNR reductions of up to $-$0.475.
\end{abstract}

\section{Introduction}
\begin{figure*}
\centering
  \includegraphics[width=0.87\textwidth, trim={0cm 25cm 0cm 0cm}, clip]{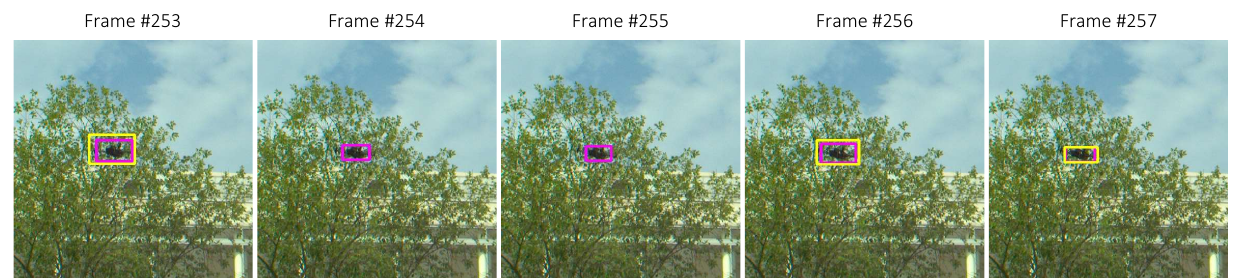}
  \caption{\label{fig:intro_example}YOLO-FEDER FusionNet~\cite{Lenhard:2024_YOLOFEDER,Lenhard:2025_YOLOFEDER} exhibits detection dropouts (yellow), producing temporally inconsistent outputs despite the drone's visibility across frames. In contrast, the detector-augmented SAMURAI maintains stable and consistent detections (magenta).}
\end{figure*}

Unmanned aerial vehicles (UAVs), in particular off-the-shelf multi-rotor drones, have become increasingly prevalent in both civilian and military domains. While they enable a wide range of applications, their misuse poses significant security risks, making reliable drone detection a critical component of modern surveillance systems~\cite{Mohsan:2023}. Recent drone detection approaches (\eg, SafeSpace MFNet~\cite{Khan:2024}, YOLOv8-MDS~\cite{Chen:2024}, YOLO-FEDER FusionNet~\cite{Lenhard:2024_YOLOFEDER,Lenhard:2025_YOLOFEDER}) achieve strong performance on individual frames, but fail to maintain consistency across time~\cite{Lenhard:2025_YOLOFEDER}. The lack of motion modeling and temporal coherence in frame-based methods leads to frequent detection dropouts (see Fig.~\ref{fig:intro_example}). While isolated dropouts are acceptable, accumulated errors limit their applicability in real-world scenarios where persistent monitoring of drones is essential.

\vspace{-0.03cm}Despite its importance, research on RGB-based drone tracking remains relatively limited~\cite{Yasmeen:2025}, with most existing approaches relying on traditional motion models --  such as Kalman filters (KFs)~\cite{Gunjal:2018} -- to propagate detections over time. In contrast, related tasks such as pedestrian or vehicle tracking have seen substantial progress, driven by foundation models with strong zero-shot generalization capabilities. Among these, \textit{SAMURAI}~\cite{Yang:2024} -- a zero-shot visual tracker with motion-aware instance-level memory -- has shown strong category-agnostic performance on standard visual object tracking (VOT) benchmarks (\eg, LaSOT~\cite{Fan:2019_LASOT}, $\text{LaSOT}_{\text{ext}}$~\cite{Fan:2020_LASOT}, or GOT-10k~\cite{Huang:2021}). In particular, it addresses fundamental VOT challenges, such as object occlusion and deformation~\cite{Zhang:2025}, which are critical to maintaining reliable trajectories in unconstrained environments. However, existing benchmarks such as LaSOT~\cite{Fan:2019_LASOT} are dominated by everyday object categories (\eg, pedestrians, vehicles, animals), with drones either entirely absent or marginally represented (\eg, 1/80 in LaSOT), without dedicated evaluation. Consequently, the applicability of such models to drone tracking remains underexplored.

To address this gap, we investigate the effectiveness of SAMURAI for drone tracking, focusing on surveillance scenarios that rely on conventional passive RGB cameras (as opposed to active pan-tilt systems~\cite{Jiang:2023}). We conduct a systematic evaluation of its performance across key sequence characteristics (\eg, environmental conditions, drone size, and temporal duration), highlighting both strengths and limitations. To quantify the sensitivity of SAMURAI to initial bounding box localization, we compare two initialization strategies: ground-truth (GT) boxes and detections from YOLO-FEDER FusionNet~\cite{Lenhard:2024_YOLOFEDER,Lenhard:2025_YOLOFEDER}, a state-of-the-art image-based drone detection model (selected for its strong performance in complex scenes). Extending this evaluation, we further introduce a detector-augmented variant of SAMURAI to mitigate initialization errors and enhance robustness, particularly in long-duration sequences. Our experiments are conducted on the publicly available DUT Anti-UAV dataset~\cite{Zhao:2022} (20 sequences, up to $\sim$2.6k frames) as well as on four custom-recorded sequences with increased temporal duration (up to 6k frames). Accordingly, we make the following key \textbf{contributions}:\vspace{0.1cm}

\begin{itemize}
    \item A comprehensive evaluation of SAMURAI for RGB-based drone tracking across diverse conditions, scales, and sequence lengths;
    \item A detector-augmented variant of SAMURAI for enhanced robustness in long-duration sequences;
    \item A new dataset comprising four long-duration drone sequences (up to 6k frames with increased scale variation), complementing DUT Anti-UAV and released publicly via Zenodo~\cite{Lenhard:2025_RealDataDrones} to foster future research. 
\end{itemize}

\noindent Together, these contributions provide the first systematic study of SAMURAI in the context of drone tracking, showing its potential and limitations for real-world surveillance applications. The remaining paper is organized as follows: Sec.~\ref{sec:related_work} reviews related work. Sec.~\ref{sec:method} provides descriptions of SAMURAI and its detector-augmented extension, and introduces the evaluation datasets and protocol. Evaluation results are presented in Sec.~\ref{sec:results}, followed by conclusions in~Sec.~\ref{sec:conclusion}. 

\section{Related Work}
\label{sec:related_work}
This section reviews related work on visual object tracking, drone detection and tracking, and public drone datasets.

\subsection{Visual Object Tracking} 
\label{subsec:vot}
Visual object tracking (VOT) is a fundamental computer vision task~\cite{Hong:2024} with broad applications in video surveillance~\cite{Abba:2024}, autonomous driving~\cite{Guo:2022}, and robotics~\cite{Gad:2022}. The goal is to estimate the trajectory of a target throughout a video sequence given its initial state, typically specified as a bounding box~\cite{Marvasti:2022}. Depending on the number of targets, VOT is commonly divided into single object tracking (SOT) and multiple object tracking (MOT)~\cite{Zhang:2025}. In SOT, the primary challenge lies in handling appearance changes, occlusion, and background clutter, whereas MOT additionally requires solving the problem of consistent data association across multiple targets~\cite{Luo:2021}.

Traditional VOT techniques rely on hand-crafted appearance and motion models, such as Kalman filters (KF)~\cite{Gunjal:2018}, optical flow~\cite{Beauchemin:1995}, or feature-flow estimation~\cite{Jin:2022}. While computationally efficient, these approaches often fail in complex real-world scenarios with significant appearance variations. The emergence of deep learning (DL) has shifted VOT from hand-crafted representations to data-driven feature learning~\cite{Zhang:2025}. Within DL-based tracking, two primary categories have been established: (i)~Siamese networks (\eg, Siam R-CNN~\cite{Voigtlaender:2019}, SiamRPN++~\cite{Li:2018}, SiamBAN~\cite{Chen:2020}, STMTrack~\cite{Fu:2021}), which formulate tracking as a similarity matching problem between a target template and a search region; and (ii)~transformer architectures (\eg, UncTrack~\cite{Yao:2025}), which leverage self-attention to capture global context and long-range dependencies~\cite{Feng:2024}.  More recently, large-scale pre-trained transformer models (foundation models) -- such as SAM 2~\cite{Ravi:2024} or SAMURAI~\cite{Yang:2024} --  have emerged as promising solutions, due to their strong zero-shot performance~\cite{Hong:2024}. SAMURAI, in particular, demonstrates enhanced feature encoding and robustness in challenging scenarios by leveraging KF-based motion modeling and motion-aware memory selection~\cite{Yang:2024}.

\subsection{Drone Detection and Tracking} 
\label{subsec:drone_detect_track}
Recent advancements in visual drone detection are largely driven by DL, with convolutional neural networks (CNNs, \eg, YOLO variants~\cite{Redmon:2015}) and transformers (\eg, Deformable DETR~\cite{Wang:2021}) constituting the architectural foundation~\cite{Bala:2025}. Generic object detection models are commonly adapted to specific application scenarios, targeting challenges such as small-drone detection~\cite{Lv:2022,Liu:2021,Chen:2024,Kim:2023,Dong:2023}, reliable discrimination from visually similar aerial objects (\eg, birds)~\cite{Singha:2021,Lv:2022}, and robustness in complex urban environments~\cite{Lv:2022,Khan:2024} -- particularly under camouflage effects~\cite{Lenhard:2024_YOLOFEDER,Lenhard:2025_YOLOFEDER}. In practice, such adaptations are typically realized either through domain-specific data curation and fine-tuning, or via architectural augmentation with task-oriented layers and modules~\cite{Zhou:2023,Lv:2022,Khan:2024}. Beyond these strategies, alternative directions explore hybrid models -- either by coupling generic detectors with complementary image-based techniques (\eg, YOLO-FEDER FusionNet~\cite{Lenhard:2024_YOLOFEDER,Lenhard:2025_YOLOFEDER}), or by integrating multi-modal data sources to exploit cross-domain information~\cite{Svanstroem:2022}. 

In contrast to detection, research on drone tracking remains comparatively scarce~\cite{Yasmeen:2025}. Whereas early correlation-filter approaches proved unstable under occlusion and appearance variations~\cite{Yasmeen:2025}, state-of-the-art drone tracking techniques primarily rely on Siamese networks and CNN-based architectures. For instance, Cheng et al.~\cite{Cheng:2022} introduce SiamAD, which extends SiamRPN++~\cite{Li:2018} with a hybrid attention mechanism, a hierarchical discriminator for confidence estimation, a YOLOv5-based re-detection module, and an adaptive template updating strategy. Wang et al.~\cite{Wang:2022} target low-altitude cluttered scenarios by incorporating attention-enhanced feature extraction, occlusion sensing, and LSTM-based trajectory prediction for drone re-localization. Unlike~\cite{Cheng:2022} and \cite{Wang:2022}, Alshaer et al.~\cite{Alshaer:2025} adopt a tracking-by-detection approach, integrating YOLO variants with KF to enhance robustness in urban drone tracking. Other trackers, including SiamFC~\cite{Bertinetto:2016}, SiamRPN++~\cite{Li:2018}, ECO~\cite{Danelljan:2017}, ATOM~\cite{Danelljan:2019}, DiMP~\cite{Bhat:2019}, TransT~\cite{Chen:2021_TransT}, SPLT~\cite{Bin:2019}, and LTMU~\cite{Dai:2020} are evaluated for drone tracking in~\cite{Zhao:2022} -- both standalone and in combination with detection models -- demonstrating promising performance. Transformer-based foundation models such as SAMURAI~\cite{Yang:2024} remain unexplored in drone tracking.

\subsection{Datasets} 
\label{subsec:drone_tracking_datasets}
The availability of annotated datasets for image-based drone detection and tracking remains limited~\cite{Yasmeen:2025,Lenhard:2024}. Existing datasets can be broadly divided into two categories: (i)~detection-oriented datasets, which predominantly consist of individual image samples, and (ii)~tracking-oriented datasets, which provide temporally ordered video data. The majority of datasets are sourced from online platforms~\cite{Aksoy:2019} or self-recorded footage~\cite{Zhao:2022}, while synthetic datasets remain rare exceptions (\eg, Sim2Air~\cite{Barisic:2022} or SynDroneVision~\cite{Lenhard:2024}). Overall, available data collections are typically problem-specific (\eg, the Drone-vs-Bird detection dataset~\cite{Coluccia:2024}) and biased toward detection. Additionally, they exhibit substantial heterogeneity in scale, resolution, sensing configurations (RGB vs. IR~\cite{Svanstroem:2021}, static~\cite{Lenhard:2024} vs. moving~\cite{Zhao:2022}, pan-tilt systems~\cite{Jiang:2023}), drone diversity, environmental settings (aerial~\cite{Barisic:2021} to urban~\cite{Lenhard:2024}), and annotation standards. Further details are provided in~\cite{Yasmeen:2025,Lenhard:2024}.

Within the domain of drone tracking, the DUT Anti-UAV~\cite{Zhao:2022} and the Anti-UAV~\cite{Jiang:2023} datasets are among the most frequently used. The Anti-UAV~\cite{Jiang:2023} dataset, collected with a pan-tilt system, is specifically designed for multimodal tracking and predominantly contains scenes with backgrounds dominated by buildings and sky. In contrast, the DUT Anti-UAV dataset, acquired with conventional RGB cameras, largely represents urban environments with greater visual complexity.

\begin{figure*}
\centering
  \includegraphics[width=0.86\textwidth, trim={0.5cm 21.1cm 0.5cm 0.05cm}, clip]{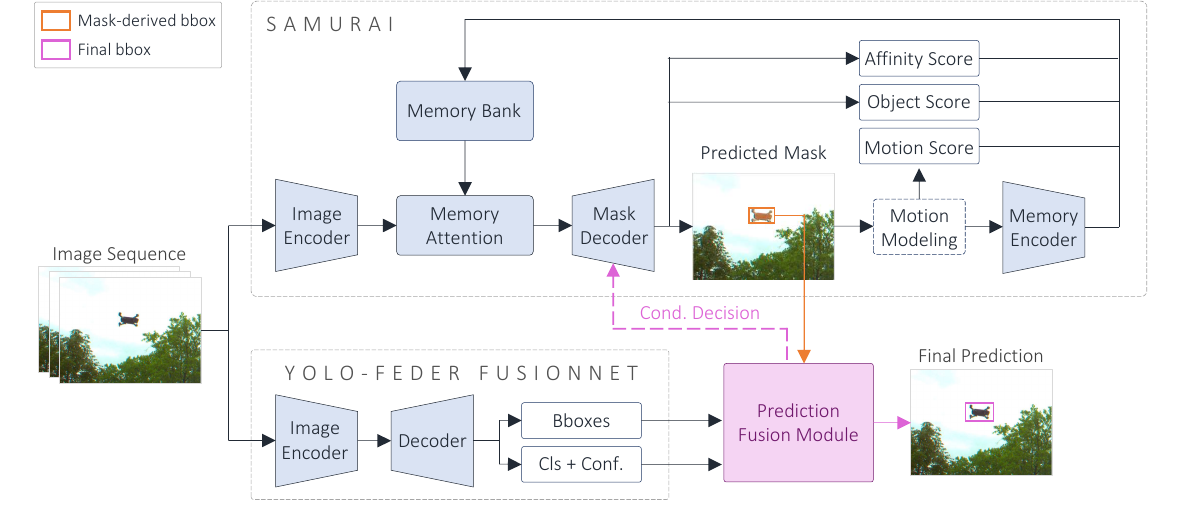}
  \caption{\label{fig:process} Schematic overview of the SAMURAI~\cite{Yang:2024} tracking pipeline extended by a detector module, specifically YOLO-FEDER FusionNet~\cite{Lenhard:2024_YOLOFEDER,Lenhard:2025_YOLOFEDER}. YOLO-FEDER FusionNet predictions are fused with mask-derived bounding boxes via a \textit{Prediction Fusion Module} to refine object localization and improve tracking consistency.}
\end{figure*}

\section{Methodology}
\label{sec:method}
This section introduces the baseline SAMURAI tracker along with our proposed detector-augmented extension, and outlines the evaluation datasets and protocol.

\subsection{SAMURAI}
\label{subsec:samurai_baseline}
SAMURAI builds upon SAM 2~\cite{Ravi:2024} and comprises five key components: (i)~an image encoder, (ii)~a mask decoder with an integrated prompt encoder, (iii)~a memory encoder, (iv)~a memory attention layer, and (v)~a memory bank (cf. Fig.~\ref{fig:process}, top). The image encoder extracts frame-level visual features, while the prompt encoder supports both sparse (points, bounding boxes) and dense (masks) inputs. In visual tracking, the target is typically initialized with the GT bounding box in the first frame, and subsequent prompts are derived from previously predicted masks. The mask decoder fuses prompt tokens with memory-conditioned image embeddings to output multiple candidate masks, each associated with a mask affinity score and a frame-level objectness score. Memory conditioning is realized by encoding selected masks into embeddings via the memory encoder, storing them in the memory bank, and retrieving them through the memory attention layer to integrate temporal context into the current prediction.

To improve the reliability of mask predictions and memory embeddings, SAMURAI incorporates motion modeling and motion-aware memory selection. For motion modeling, a linear KF is employed to predict target dynamics. For each candidate mask generated by the decoder, a KF-IoU score is computed between the predicted state and the mask-derived bounding box. The final mask is selected by maximizing a weighted combination of the KF-IoU and mask affinity scores. The filter is updated in a predict-correct cycle with gated influence to ensure stability under occlusions. The KF-IoU score further supports the motion-aware memory selection, where candidate frames are evaluated by (i)~mask affinity, (ii)~objectness, and (iii)~motion consistency (KF-IoU). Only the highest-scoring frames are retained in the memory bank, ensuring that memory attention conditions predictions on reliable contextual features while mitigating error accumulation. For further details on SAMURAI and SAM 2, refer to \cite{Yang:2024} and \cite{Ravi:2024}.

\subsection{Detector-Augmented SAMURAI}
\label{subsec:samurai_extension}
To enhance tracking robustness and mitigate error accumulation in long-duration sequences -- an inherent challenge for SAMURAI~\cite{Yang:2024,Ravi:2024} -- we introduce a detector-augmented extension. Specifically, we integrate SAMURAI with the task-specific detection model YOLO-FEDER FusionNet~\cite{Lenhard:2024_YOLOFEDER,Lenhard:2025_YOLOFEDER}. The outputs from both models -- YOLO-FEDER FusionNet’s bounding box predictions with confidence estimates and SAMURAI’s mask-derived bounding box -- are fused within the \textit{Prediction Fusion Module} (see Fig.~\ref{fig:process}, magenta), which performs conditional integration to enable detector-guided re-initialization and correction. 

The following sections provide a brief overview of YOLO-FEDER FusionNet, followed by a detailed description of the Prediction Fusion Module, the principal mechanism for coupling detection with tracking.
\vspace{-0.45cm}
\paragraph{YOLO-FEDER FusionNet.} YOLO-FEDER FusionNet \cite{Lenhard:2024_YOLOFEDER,Lenhard:2025_YOLOFEDER} is a DL model developed for drone detection in visually complex environments, where standard detectors such as YOLOv5~\cite{Ultralytics_YOLOv5} or YOLOv8~\cite{Ultralytics_YOLOv8} exhibit degraded performance (\eg, under camouflage effects or in low-contrast conditions). The architecture adopts a dual-backbone fusion design, coupling a YOLOv8l backbone with the camouflage object detection model FEDER~\cite{He:2023} to exploit complementary feature representations. Feature maps from both backbones are fused within a shared neck architecture, followed by a YOLOv8-inspired detection head. Trained on a combination of synthetic and real-world data -- specifically SynDroneVision~\cite{Lenhard:2024} and DUT Anti-UAV~\cite{Zhao:2022} (detection subset) -- YOLO-FEDER FusionNet demonstrates strong real-time performance in visually complex environments, processing frames in just 12.4~ms on an NVIDIA A100. Therefore, the pretrained model is employed as the detector component within the detector-augmented extension of SAMURAI. (Note that the DUT Anti-UAV detection and tracking datasets are distinct subsets; see Sec.~\ref{subsec:datasets}.) Further details on YOLO-FEDER FusionNet are provided in~\cite{Lenhard:2024_YOLOFEDER,Lenhard:2025_YOLOFEDER}.

\vspace{-0.4cm}
\paragraph{Prediction Fusion Module.} The proposed Prediction Fusion Module integrates outputs from both the SAMURAI tracker and the YOLO-FEDER FusionNet detector. Its primary role is to enhance tracking performance by leveraging detector cues for re-localizing lost targets, refining bounding boxes, and providing periodic corrections.

The module's core mechanism is continuous prompting (beyond the initial first-frame bounding box prompt, cf. Sec.~\ref{subsec:samurai_baseline}), in which reliable detector predictions are periodically injected into the tracker to mitigate drift and reinforce stability. A new bounding box prompt is introduced whenever a YOLO-FEDER FusionNet detection is available and deemed reliable -- based on high confidence ($>0.75$), strong spatial alignment with the SAMURAI estimate, or proximity to recent trajectory history. Here, strong spatial alignment is defined as high overlap ($>0.7$) between SAMURAI and YOLO-FEDER FusionNet predictions, measured by the complete intersection over union (CIoU)~\cite{Zheng:2020_1,Zheng:2020_2}. The proximity to recent trajectory history is determined by comparing the current YOLO-FEDER FusionNet detection against prior bounding boxes within a fixed temporal window. A detection is considered in proximity if it either overlaps sufficiently with the most recent prediction ($\text{IoU} > 0.8$) or its normalized center distance is below a predefined threshold. 

In our experiments, we adopt a temporal window of 10 frames and set the center-distance threshold to 0.05 (5\% of the frame dimensions), which yields a robust trade-off between tolerance to minor localization noise and sensitivity to significant trajectory deviations. To avoid error accumulation during long-term propagation, we enforce bounding-box prompting at least once every 30 frames ($\sim$1s at 30 fps), conditioned on the availability of a reliable detection. In the absence of YOLO-FEDER detections, the module defaults to SAMURAI predictions for uninterrupted propagation.

Finally, if the SAMURAI bounding box is fully enclosed by the YOLO-FEDER FusionNet detection, the final estimate is given by the arithmetic mean of their bounding-box coordinates. This adjustment is motivated by empirical observations that YOLO-FEDER FusionNet detections are generally more reliable, given that the detector is specifically trained for drone detection, whereas SAMURAI operates in a zero-shot mode. Since the Prediction Fusion Module is modular and detector-agnostic -- allowing YOLO-FEDER FusionNet to be replaced with arbitrary object detectors (\eg, cf.~\cite{Yun:2025}) -- the averaging adjustment might require adaptation to the detector's reliability characteristics.

\subsection{Datasets}
\label{subsec:datasets}
SAMURAI and its detector-augmented extension are evaluated using both a publicly available dataset and custom recordings. Specifically, we employ the DUT Anti-UAV tracking dataset~\cite{Zhao:2022} and complement it with four long-duration sequences from our custom datasets R1 and R2, designed to better capture the challenges of extended tracking. Tab.~\ref{tab:data_overview} provides an overview of all sequences.
\vspace{-0.4cm}
\paragraph{DUT Anti-UAV~\cite{Zhao:2022}.} The DUT Anti-UAV dataset (tracking subset) contains 20 outdoor video sequences, each depicting a single drone. The backgrounds range from clear blue skies to cluttered tree lines with low target-background contrast. While some sequences are visually challenging, most provide comparatively favorable tracking conditions, with the drone either at a relatively large scale or clearly distinguishable due to strong foreground-background contrast. The sequences are recorded at resolutions of 1280$\times$720 or 1920$\times$1080 pixels (see Tab.~\ref{tab:data_overview}) under dynamic camera setups, with background variations induced by camera motion. They include different drone models, typically localized in the lower two-thirds of the field of view (FOV), with object area ratios between 2.7e-04 and 0.045 (mean: 0.0031). Sequence lengths vary from 83 to 2635 frames. 

Note that the detection subset (used for training YOLO-FEDER FusionNet, cf. Sec.~\ref{subsec:samurai_extension}) and the tracking subset of DUT Anti-UAV do not contain overlapping frames, even though some sequences may stem from the same recording environments. Further details can be found in~\cite{Zhao:2022} and the supplementary material (Supp.).

\vspace{-0.4cm}
\paragraph{Custom Data.} The custom datasets R1 and R2 (cf. Fig.~\ref{fig:intro_example}) are collected using a ground-mounted Basler acA200-165c camera system in an urban area characterized by medium-density vegetation and medium-height buildings. R1 is recorded with an 8 mm wide-angle lens, providing a broad FOV, while R2 is captured with a 25 mm lens, offering a closer perspective. Each dataset comprises two sequences recorded at a resolution of 2040$\times$1086 pixels (see Tab.~\ref{tab:data_overview}), with a frame rate of 60 fps. A single Yuneec Mantis G is used for all recordings. The drone appears at a small object scale, exhibits pronounced scale variation, and frequently leaves the FOV (up to 100$+$ consecutive frames). Sequence lengths range from 1484 to 6327 frames, offering extended temporal coverage for long-term tracking evaluation. Further details are provided in the supplementary material.

\begin{table}[t!]
\caption{Overview of image sequences from DUT Anti-UAV~\cite{Zhao:2022} and custom datasets R1 and R2. The table reports sequence length, resolution, and drone visibility (\textit{Leaves FOV}, last column).}
\label{tab:data_overview}
\footnotesize
  \centering
  \begin{tabular}{@{}|c|l|c|c|c|@{}}
    \hline
    Dataset & \multicolumn{1}{c|}{Seq.} & No. & Resolution & Leaves FOV\\
    & & Frames & & (yes~/~no)\\\hline
    \multirow{19}{*}{DUT Anti-}           & video01 & 1050 & 1920$\times$1080 & no\\\cline{2-5}
    \multirow{19}{*}{UAV~\cite{Zhao:2022}}& video02 & 83   & 1280$\times$720 & no\\\cline{2-5}
    & video03 & 100  & 1280$\times$720  & no\\\cline{2-5}
    & video04 & 341  & 1920$\times$1080 & no\\\cline{2-5}
    & video05 & 450  & 1280$\times$720  & no\\\cline{2-5}
    & video06 & 200  & 1920$\times$1080 & no\\\cline{2-5}
    & video07 & 2480 & 1280$\times$720  & yes\\\cline{2-5}
    & video08 & 2305 & 1280$\times$720  & yes\\\cline{2-5}
    & video09 & 2500 & 1920$\times$1080 & yes\\\cline{2-5}
    & video10 & 2635 & 1920$\times$1080 & yes\\\cline{2-5}
    & video11 & 1000 & 1920$\times$1080 & no\\\cline{2-5}
    & video12 & 1485 & 1920$\times$1080 & yes\\\cline{2-5}
    & video13 & 1915 & 1920$\times$1080 & no\\\cline{2-5}
    & video14 & 590  & 1920$\times$1080 & no\\\cline{2-5}
    & video15 & 1350 & 1920$\times$1080 & yes\\\cline{2-5}
    & video16 & 1285 & 1920$\times$1080 & no\\\cline{2-5}
    & video17 & 780  & 1920$\times$1080 & no\\\cline{2-5}
    & video18 & 1320 & 1920$\times$1080 & yes\\\cline{2-5}
    & video19 & 1300 & 1920$\times$1080 & no\\\cline{2-5}
    & video20 & 1635 & 1920$\times$1080 & no\\\hline
    
    \multirow{2}{*}{R1} & POS3 & 6213 & 2040$\times$1086 & yes\\\cline{2-5}
    & POS7 & 6327 & 2040$\times$1086 & yes\\\hline
    \multirow{2}{*}{R2} & POS3 & 1484 & 2040$\times$1086 & yes\\\cline{2-5}
    & POS7 & 4908 & 2040$\times$1086 & yes\\\hline
  \end{tabular}
\end{table}

\begin{table*}
\centering\footnotesize
\caption{Performance of SAMURAI and its detector-augmented extension (\ding{51}, 3rd column). Metrics are aggregated by averaging per-sequence results within each dataset. \textit{GT} (2nd column) denotes initialization with ground truth, while \textit{D} denotes detector-based initialization using the first YOLO-FEDER FusionNet prediction as the bounding-box prompt. Best results are highlighted in \textbf{bold}.}
\label{tab:SAMURAI_notSeqDep}
  \begin{tabular}{@{}|l|c|c|ccc|ccc|c|c|@{}}
    \multicolumn{1}{c}{ } & \multicolumn{1}{c}{ } & \multicolumn{1}{c}{ }& \multicolumn{3}{c}{Tracking Metrics} & \multicolumn{5}{c}{Detection Metrics}\\
    \hline
    Dataset & Init. & Detector & S~$\uparrow$ & P~$\uparrow$ & $\text{P}_{\text{norm}}$~$\uparrow$& \multicolumn{3}{c|}{mAP~$\uparrow$} & FNR~$\downarrow$ & FDR~$\downarrow$ \\
    & Method & Augmentation & & & & {\small @0.25} & {\small @0.5} & {\small @0.5-0.95} & & \\\hline
    \multirow{3}{*}{DUT Anti-UAV~\cite{Zhao:2022}} & GT & -- & 0.663 & 0.888 & \textbf{0.973} & 0.958 & 0.720 & 0.398 & 0.023 & 0.031\\
    & D & -- & 0.614 & 0.842 & 0.925 & 0.909 & 0.641 & 0.354 & 0.072 & 0.080 \\
    & D & \ding{51} & \textbf{0.725} & \textbf{0.924} & 0.971 & \textbf{0.976} & \textbf{0.865} & \textbf{0.490} & \textbf{0.012} & \textbf{0.017} \\\hline
    
    \multirow{3}{*}{R1} & GT & -- & 0.242 & 0.376 & 0.402 & 0.689 & 0.494 & 0.154 & 0.591 & 0.005\\
    & D & --  & 0.343 & 0.565 & 0.608 & 0.785 & 0.496 & 0.146 & 0.384 & 0.008\\
    & D & \ding{51} & \textbf{0.635} & \textbf{0.818} & \textbf{0.869} & \textbf{0.939} & \textbf{0.853} & \textbf{0.426} & \textbf{0.116} & \textbf{0.004}\\\hline
    
    \multirow{3}{*}{R2} & GT & -- & 0.378 & 0.593 & 0.821 & 0.737 & 0.399 & 0.124 & 0.312 & 0.316\\
    & D & --  & 0.374 & 0.595 & 0.815 &  0.738 & 0.387 & 0.123 & 0.315 & 0.319 \\
    & D & \ding{51} & \textbf{0.569} & \textbf{0.832} & \textbf{0.927} & \textbf{0.959} & \textbf{0.644} & \textbf{0.264} & \textbf{0.066} & \textbf{0.017}\\\hline
  \end{tabular}
\end{table*}

\subsection{Evaluation Protocol}
\label{subsec:metrics}
We evaluate SAMURAI under two distinct initialization strategies. The first adheres to the standard VOT protocol, in which the tracker is initialized using the GT bounding box of the first frame. To reflect practical deployment conditions without GT annotations, the second strategy employs tracker initialization using the first bounding box predicted by the detection model. Evaluation of the detector-augmented version of SAMURAI focuses exclusively on the second initialization strategy. To quantify the performance of SAMURAI and assess the additional benefits provided by its detector-augmented extension, we employ the following VOT metrics:

\begin{itemize}
    \item \textbf{Success Rate (S)} -- percentage of frames where the predicted box $B^{pred}$ overlaps with the ground truth $B^{gt}$ by at least $\tau \in [0,1]$. Thus, tracking at frame $t$ is considered successful if $\mathrm{IoU}(B_t^{pred}, B_t^{gt}) \geq \tau$. Varying $\tau$ from 0 to 1 in increments of 0.01 yields the success curve; the final score $S$ corresponds to the area under the curve (AUC).
    
    \item \textbf{Precision (P)} -- percentage of frames in which the Euclidean distance between the predicted target center $c^{pred}$ and the GT center $c^{gt}$ is below a threshold $\delta$. Therefore, the prediction at frame $t$ is considered correct if $e_t = \lVert c_t^{pred} - c_t^{gt} \rVert_2 \leq \delta$. In practice, $\delta$ is varied from 0 to 50 pixels with a step size of 1 to obtain the precision curve~\cite{Zhao:2022}, and the final score P is given by the AUC.

    \item \textbf{Normalized Precision ($\text{P}_{\text{norm}}$)} -- resolution-invariant extension of precision obtained by normalizing the center error $e_t$ by the image diagonal $d$. Thus, tracking in frame $t$ is correct if $\tilde{e}_t = e_t / d \leq \tilde{\delta}$. Here, $\tilde{\delta}$ is typically varied from 0 to 0.5 in increments of 0.01 to generate the corresponding curve~\cite{Zhao:2022}.
\end{itemize}

\noindent For direct comparison with frame-by-frame detection (\eg, via YOLO-FEDER FusionNet), tracking outputs are also interpreted as per-frame detections and evaluated using established object detection metrics. Specifically, we report mean average precision (mAP) at IoU thresholds 0.25 and 0.5, mAP averaged over 0.5-0.95, false negative rate (FNR), and false discovery rate (FDR) (see~\cite{Bala:2025} for details).

\section{Results}
\label{sec:results}
In this section, we present the evaluation results for SAMURAI under zero-shot conditions and its detector-augmented extension. The Supp. provides further details on sequence-level analyses and supplementary visual examples.

\subsection{Zero-Shot SAMURAI}
\label{subsec:results_SAMURAI}
\begin{figure*}
\centering
  \includegraphics[width=0.87\textwidth, trim={0.2cm 23cm 0.2cm 0.1cm}, clip]{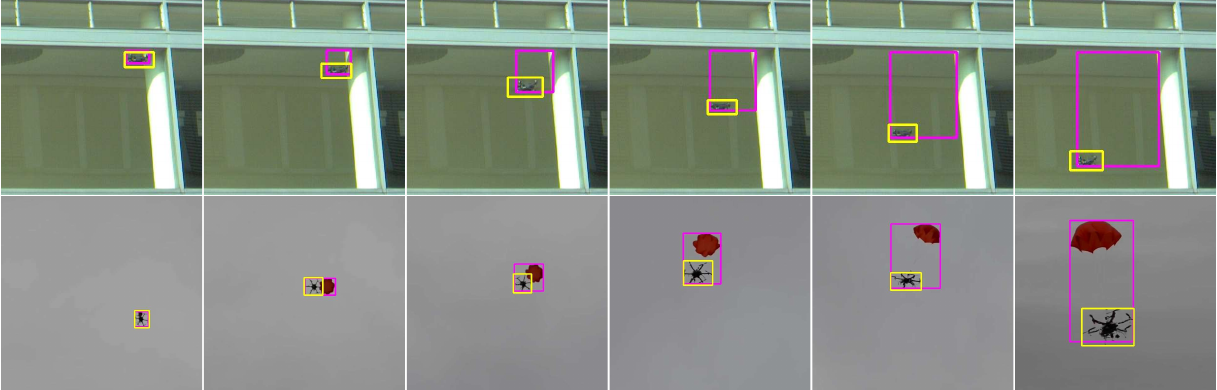}
  \caption{\label{fig:examples_1} Limitations of SAMURAI when conditioned only on the first-frame bounding box. Yellow boxes denote YOLO-FEDER FusionNet detections, while magenta boxes indicate SAMURAI predictions. Without detector guidance, SAMURAI seems to drift to irrelevant regions or include background structures, leading to errors that propagate across frames.}
\end{figure*}

\paragraph{Performance on DUT Anti-UAV.} Evaluating SAMURAI with first-frame GT bounding box initialization demonstrates strong zero-shot performance on the DUT Anti-UAV dataset~\cite{Zhao:2022}. On average, the tracker achieves a success rate of 0.663 (cf. Tab.~\ref{tab:SAMURAI_notSeqDep}), with values ranging from 0.412 to 0.893 across sequences (cf. Tab.~II, Supp.). Bounding box center precision remains high, with mean values of 0.824 (P) and 0.973 ($\text{P}_{\text{norm}}$). The only exception is \textit{video04}, where appearance variations induced by external attachments yield erroneous segmentation masks and thus inaccurate mask-derived bounding boxes (see Fig.~\ref{fig:examples_1}, bottom row). Detection metrics further show strong accuracy at coarse IoU thresholds but a decline under stricter criteria (cf. Tab.~\ref{tab:SAMURAI_notSeqDep}, mAP@0.25 vs. mAP@0.5-0.95), while FNR and FDR remain low. Compared to the trackers in~\cite{Zhao:2022}, SAMURAI improves by 0.055~(S), 0.056~(P), and 0.115~($\text{P}_{\text{norm}}$) over the best-performing model (cf. Tab.~I, Supp.).

With detector-based initialization, comparable trends are observed, albeit with an average metric degradation of 0.044 to 0.079 relative to GT initialization. An exception is \textit{video05}, where erroneous detector initialization causes the tracker to follow an unrelated object without recovery, resulting in severely degraded performance (cf. Tab.~II and Fig.~II, bottom row, Supp.).

\begin{figure*}
\centering
  \includegraphics[width=0.87\textwidth, trim={0.2cm 21.5cm 0.2cm 4.2cm}, clip]{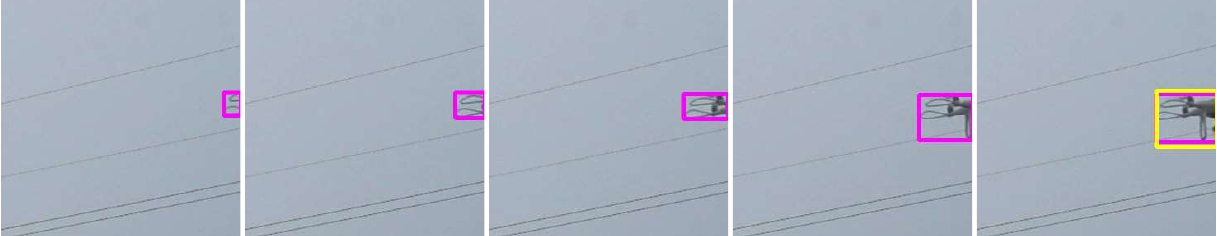}
  \caption{\label{fig:examples_FPs} Comparison of GT annotations (yellow) and detector-augmented SAMURAI predictions (magenta). While SAMURAI tracks the drone continuously from its first appearance (rightmost), GT labels (leftmost) often begin later, resulting in incomplete temporal coverage.}
\end{figure*}

\vspace{-0.4cm}
\paragraph{Performance on Custom Data.} Compared to DUT Anti-UAV, GT initialization of SAMURAI yields markedly lower performance on R1 and R2, with average success and precision values below 0.378 and 0.592, respectively (cf. Tab.~\ref{tab:SAMURAI_notSeqDep}). On R1, it further exhibits low normalized precision. Beyond the moderate localization accuracy (cf. Tab.~\ref{tab:SAMURAI_notSeqDep}, mAP values), detection metrics expose high average FNRs of 0.591 (R1) and 0.312 (R2), along with a notably high FDR on R2. However, performance varies considerably across individual sequences. For instance, the FNR on R2 drops to 0.009 for \textit{POS7}, but rises to 0.614 for \textit{POS3} (cf. Tab.~III, Supp.), with \textit{POS3} also exhibiting a high FDR. The elevated FNR and FDR in \textit{POS3} stem from the tracker drifting to background structures instead of the target, leading to systematic missed detections and false alarms (see Fig.~\ref{fig:examples_1}, top row).

Initializing SAMURAI with the first detector-derived bounding box has negligible impact on R2, with tracking and detection metrics closely matching those from GT initialization. In contrast, on R1 -- particularly for \textit{POS3} -- detector-based initialization yields improvements across all metrics, most notably reducing the average FNR from 0.591 to 0.384. However, a notable limitation lies in the occasionally unstable (often oversized) predictions that occur when the drone leaves and reenters the FOV (cf. Fig.~IV, Supp.).

\subsection{Detector-Augmented SAMURAI}
\label{subsec:results_SAMURAI_enhanced}
\paragraph{Performance on DUT Anti-UAV.} SAMURAI's detector-augmented extension improves both tracking and detection metrics over first-frame-only initialization (cf. Tab.~\ref{tab:SAMURAI_notSeqDep}), with the strongest gains on \textit{video04} and \textit{video05} (see Tab.~II, Supp.). Continuous conditioning through detector-derived bounding boxes enables the tracker to recover under appearance changes (\textit{video04}), while also mitigating errors from inaccurate initialization by re-localizing the target (\textit{video05}). A corresponding visualization is provided in the Supp. (Fig.~I). While these results demonstrate clear benefits, some sequences still exhibit elevated FDRs. Closer analysis reveals that these cases occur when the drone partially or fully leaves the FOV. The tracker follows the target until it fully disappears and resumes immediately upon re-entry, whereas the GT often terminates earlier or begins later, as shown in Fig.~\ref{fig:examples_FPs}. This mismatch leads to apparent FPs during evaluation that stem from annotation inconsistencies rather than tracking failures. Compared to the best-performing tracking-with-detection combinations reported in~\cite{Zhao:2022}, the detector-augmented SAMURAI achieves improved performance, especially w.r.t. success rate and normalized precision (see Tab.~\ref{tab:SAMURAI_track+detect}).

\begin{table}[t!]
\footnotesize
\caption{\label{tab:SAMURAI_track+detect}Comparison of detector-augmented SAMURAI with the best-performing tracking-with-detection combinations from~\cite{Zhao:2022} on DUT Anti-UAV. Detector notation: \textless model\textgreater-\textless backbone\textgreater.}
  \begin{tabular}{@{}|l|l|ccc|@{}}\hline
    Tracker & Detector & S~$\uparrow$ & P~$\uparrow$ & $\text{P}_{\text{norm}}$~$\uparrow$\\\hline
    TransT~\cite{Chen:2021_TransT} & CRCNN-ResNet50 & 0.624 & 0.888 & 0.808\\\hline
    DiMP~\cite{Bhat:2019} & FRCNN-ResNet50 & 0.657	& 0.949	& 0.856\\\hline
    ATOM~\cite{Danelljan:2019} & FRCNN-ResNet18 & 0.635 & 0.936 & 0.828\\\hline
    SiamFC~\cite{Bertinetto:2016} & \multirow{5}{*}{FRCNN-VGG16} & 0.615 & 0.943 & 0.811\\\cline{1-1}\cline{3-5}
    ECO~\cite{Danelljan:2017} & & 0.620 & 0.954 & 0.821\\\cline{1-1}\cline{3-5}
    SPLT~\cite{Bin:2019} & & 0.553	& 0.875	& 0.783\\\cline{1-1}\cline{3-5}
    SiamRPN++~\cite{Li:2018} & &  0.612 & 0.881 & 0.797\\\cline{1-1}\cline{3-5}
    LTMU~\cite{Dai:2020} & & 0.664	& \textbf{0.961} & 0.865\\\hline
    \multirow{2}{*}{\textbf{SAMURAI}~\cite{Yang:2024}} & \textbf{YOLO-FEDER} & \multirow{2}{*}{\textbf{0.725}} & \multirow{2}{*}{0.924} & \multirow{2}{*}{\textbf{0.971}}\\
    & \textbf{FusionNet}  & & & \\\hline
    \multicolumn{5}{l}{FRCNN = Faster R-CNN, \hspace{0.1cm}CRCNN = Cascade R-CNN}\\
  \end{tabular}
\end{table}

\begin{figure*}
\centering
  \includegraphics[width=0.86\textwidth, trim={0.3cm 16.8cm 0.3cm 0.2cm}, clip]{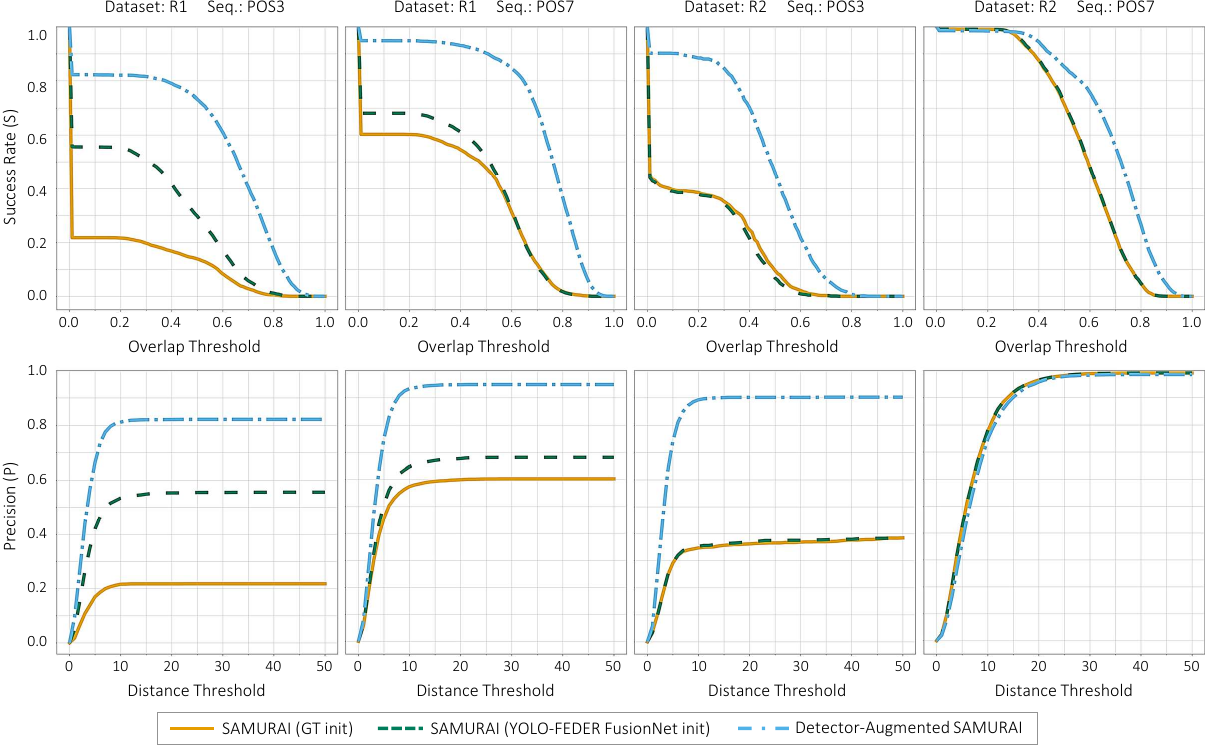}
  \caption{\label{fig:success_rate_R1R2} Success rate (top) and precision (bottom) curves across sequences from datasets R1 and R2, comparing SAMURAI with GT initialization, SAMURAI with YOLO-FEDER FusionNet initialization, and the detector-augmented SAMURAI.}
\end{figure*}
\vspace{-0.4cm}
\paragraph{Performance on Custom Data.} Evaluations on datasets R1 and R2 show that integrating YOLO-FEDER FusionNet detections with SAMURAI yields significant performance improvements, particularly in long-duration sequences. Compared to first-frame-only detector-based initialization, the proposed tracking strategy yields absolute success rate gains of 0.292 on R1 (0.343~vs.~0.635) and 0.195 on R2 (0.374~vs.~0.569). This is also reflected in the sequence-level success curves (Fig.~\ref{fig:success_rate_R1R2}, top). Similar improvements are observed across (normalized) precision and detection metrics (cf. Tab.~\ref{tab:SAMURAI_notSeqDep} and Fig.~\ref{fig:success_rate_R1R2}, bottom). Particularly notable is the reduction in FNRs, with decreases of 0.475 (R1) and 0.246 (R2). Despite the substantial reduction, the detector-augmented SAMURAI occasionally still struggles with very small drones (cf. Fig.~V, Supp.)

\subsection{Discussion}
\label{subsec:discussion}
Overall, SAMURAI demonstrates promising zero-shot performance, highlighting the potential of transformer-based foundation models for drone tracking even in the absence of task-specific training -- particularly under favorable tracking conditions. Moreover, it surpasses previously reported non-transformer-based trackers by a significant margin~\cite{Zhao:2022}. However, several limitations arise in more complex or realistic deployment scenarios: (1)~SAMURAI is highly dependent on accurate initialization: while GT boxes provide a clean start in benchmarks, real-world detectors may produce imperfect or erroneous inputs from which it cannot recover. (2)~Substantial appearance changes often result in inconsistent bounding box predictions. Although such flexibility may be beneficial in domains with high intra-class variation (\eg, person tracking), it introduces instability in drone tracking, where the target’s appearance is relatively constrained. (3)~Performance degrades in long-duration sequences -- likely tied to limitations of the underlying SAM 2 architecture~\cite{Ravi:2024} -- as well as difficulties with target re-identification after temporary disappearances. 

The detector-augmented extension of SAMURAI directly addresses these limitations. By leveraging continuous detector cues, it reduces sensitivity to imperfect initialization, stabilizes predictions under perturbations (\eg, externally attached parachutes, Fig.~\ref{fig:examples_1}), and provides reliable re-localization in long-duration sequences. Consequently, it proves more robust in realistic deployment settings, where detection noise, target variability, and long-duration tracking remain key challenges. Nevertheless, the approach remains sensitive to detection quality and fusion parameters, which can be influenced by factors such as frame rate and drone velocity. Furthermore, robust tracking of very small drones in complex scenes is still a challenge.

\section{\label{sec:conclusion}Conclusion}
In this work, we provided the first systematic evaluation of SAMURAI for drone tracking, introduced a detector-augmented extension that enhanced robustness under realistic deployment conditions, and released a new long-duration dataset to support future research. However, sensitivity to detector quality and reliable tracking of small targets in complex environments remain key bottlenecks.

{
    \small
    \bibliographystyle{ieeenat_fullname}
    \bibliography{main}

\begin{thebibliography}{62}
\providecommand{\natexlab}[1]{#1}
\providecommand{\url}[1]{\texttt{#1}}
\expandafter\ifx\csname urlstyle\endcsname\relax
  \providecommand{\doi}[1]{doi: #1}\else
  \providecommand{\doi}{doi: \begingroup \urlstyle{rm}\Url}\fi

\bibitem[Abba et~al.(2024)Abba, Bizi, Lee, et~al.]{Abba:2024}
Sani Abba, Ali~Mohammed Bizi, Jeong-A Lee, et~al.
\newblock {Real-time Object Detection, Tracking, and Monitoring Framework for
  Security Surveillance Systems}.
\newblock \emph{Heliyon}, 10\penalty0 (15):\penalty0 e34922, 2024.

\bibitem[Aksoy et~al.(2019)Aksoy, Orak, and \"{O}zkan]{Aksoy:2019}
Mehmet~\c{C}a\u{g}ri Aksoy, Alp~Sezer Orak, and Hasan Mertcan~others \"{O}zkan.
\newblock {Drone Dataset: Amateur Unmanned Air Vehicle Detection}.
\newblock \emph{Mendeley Data}, V4, 2019.

\bibitem[Alshaer et~al.(2025)Alshaer, Abdelfatah, Ismail, et~al.]{Alshaer:2025}
Nancy Alshaer, Reham Abdelfatah, Tawfik Ismail, et~al.
\newblock {Vision-Based UAV Detection and Tracking Using Deep Learning and
  Kalman Filter}.
\newblock \emph{Comput. Intell.}, 41\penalty0 (1):\penalty0 e70026, 2025.

\bibitem[Bala et~al.(2025)Bala, Muqaibel, Iqbal, et~al.]{Bala:2025}
Abubakar Bala, Ali~H. Muqaibel, Naveed Iqbal, et~al.
\newblock {Machine Learning for Drone Detection from Images: A Review of
  Techniques and Challenges}.
\newblock \emph{Neurocomputing}, 635:\penalty0 129823, 2025.

\bibitem[Barisic et~al.(2021)Barisic, Petric, and Bogdan]{Barisic:2021}
Antonella Barisic, Frano Petric, and Stjepan Bogdan.
\newblock {Brain over Brawn: Using a Stereo Camera to Detect, Track, and
  Intercept a Faster UAV by Reconstructing the Intruder's Trajectory}.
\newblock \emph{Field Robot.}, 2:\penalty0 222--240, 2021.

\bibitem[Barisic et~al.(2022)Barisic, Petric, and Bogdan]{Barisic:2022}
Antonella Barisic, Frano Petric, and Stjepan Bogdan.
\newblock {Sim2Air - Synthetic Aerial Dataset for UAV Monitoring}.
\newblock \emph{IEEE Robot. Autom. Lett.}, 7\penalty0 (2):\penalty0 3757--3764,
  2022.

\bibitem[Beauchemin and Barron(1995)]{Beauchemin:1995}
Steven~S. Beauchemin and John~L. Barron.
\newblock {The Computation of Optical Flow}.
\newblock \emph{ACM Comput. Surv.}, 27\penalty0 (3):\penalty0 433--466, 1995.

\bibitem[Bertinetto et~al.(2016)Bertinetto, Valmadre, Henriques,
  et~al.]{Bertinetto:2016}
Luca Bertinetto, Jack Valmadre, Jo{\~a}o~F. Henriques, et~al.
\newblock {Fully-Convolutional Siamese Networks for Object Tracking}.
\newblock In \emph{Eur. Conf. Comput. Vis. Workshops}, pages 850--865, 2016.

\bibitem[Bhat et~al.(2019)Bhat, Danelljan, Van~Gool, et~al.]{Bhat:2019}
Goutam Bhat, Martin Danelljan, Luc Van~Gool, et~al.
\newblock {Learning Discriminative Model Prediction for Tracking}.
\newblock In \emph{IEEE/CVF Int. Conf. Comput. Vis.}, pages 6181--6190, 2019.

\bibitem[Chen et~al.(2024)Chen, Zheng, Sun, et~al.]{Chen:2024}
Mingxi Chen, Zhen Zheng, Haoran Sun, et~al.
\newblock {YOLOv8-MDS: A YOLOv8-Based Multi-Distance Scale Drone Detection
  Network}.
\newblock \emph{J. Phys.: Conf. Ser.}, 2891\penalty0 (15):\penalty0 152008,
  2024.

\bibitem[Chen et~al.(2021)Chen, Yan, Zhu, et~al.]{Chen:2021_TransT}
Xin Chen, Bin Yan, Jiawen Zhu, et~al.
\newblock {Transformer Tracking}.
\newblock In \emph{IEEE/CVF Conf. Comput. Vis. Pattern Recog.}, pages
  8122--8131, 2021.

\bibitem[Chen et~al.(2020)Chen, Zhong, Li, et~al.]{Chen:2020}
Zedu Chen, Bineng Zhong, Guorong Li, et~al.
\newblock {Siamese Box Adaptive Network for Visual Tracking}.
\newblock \emph{IEEE/CVF Conf. Comput. Vis. Pattern Recog.}, pages 6667--6676,
  2020.

\bibitem[Cheng et~al.(2022)Cheng, Liang, Peng, et~al.]{Cheng:2022}
Feng Cheng, Zhibo Liang, Gaoliang Peng, et~al.
\newblock {An Anti-UAV Long-Term Tracking Method with Hybrid Attention
  Mechanism and Hierarchical Discriminator}.
\newblock \emph{Sensors}, 22\penalty0 (10), 2022.

\bibitem[Coluccia et~al.(2024)Coluccia, Fascista, Sommer, Schumann, Dimou, and
  Zarpalas]{Coluccia:2024}
Angelo Coluccia, Alessio Fascista, Lars Sommer, Arne Schumann, Anastasios
  Dimou, and Dimitrios Zarpalas.
\newblock {The Drone-vs-Bird Detection Grand Challenge at ICASSP 2023: A Review
  of Methods and Results}.
\newblock \emph{IEEE Open Journal of Signal Processing}, 5:\penalty0 766--779,
  2024.

\bibitem[Dai et~al.(2020)Dai, Zhang, Wang, et~al.]{Dai:2020}
Kenan Dai, Yunhua Zhang, Dong Wang, et~al.
\newblock {High-Performance Long-Term Tracking With Meta-Updater}.
\newblock In \emph{IEEE/CVF Conf. Comput. Vis. Pattern Recog.}, pages
  6297--6306, 2020.

\bibitem[Danelljan et~al.(2017)Danelljan, Bhat, Khan, et~al.]{Danelljan:2017}
Martin Danelljan, Goutam Bhat, Fahad~Shahbaz Khan, et~al.
\newblock {ECO: Efficient Convolution Operators for Tracking}.
\newblock In \emph{IEEE/CVF Conf. Comput. Vis. Pattern Recog.}, pages
  6931--6939, 2017.

\bibitem[Danelljan et~al.(2019)Danelljan, Bhat, Khan, et~al.]{Danelljan:2019}
Martin Danelljan, Goutam Bhat, Fahad~Shahbaz Khan, et~al.
\newblock {ATOM: Accurate Tracking by Overlap Maximization}.
\newblock In \emph{IEEE/CVF Conf. Comput. Vis. Pattern Recog.}, pages
  4655--4664, 2019.

\bibitem[Dong et~al.(2023)Dong, Wang, Lu, et~al.]{Dong:2023}
Pengcheng Dong, Chuntao Wang, Zhenyong Lu, et~al.
\newblock {S-Feature Pyramid Network and Attention Model for Drone Detection}.
\newblock In \emph{IEEE Int. Conf. Acoust., Speech, Signal Process.}, pages
  1--2, 2023.

\bibitem[Fan et~al.(2019)Fan, Lin, Yang, et~al.]{Fan:2019_LASOT}
Heng Fan, Liting Lin, Fan Yang, et~al.
\newblock {LaSOT: A High-Quality Benchmark for Large-Scale Single Object
  Tracking}.
\newblock In \emph{IEEE/CVF Conf. Comput. Vis. Pattern Recog.}, 2019.

\bibitem[Fan et~al.(2020)Fan, Bai, Lin, et~al.]{Fan:2020_LASOT}
Heng Fan, Hexin Bai, Liting Lin, et~al.
\newblock {LaSOT: A High-quality Large-scale Single Object Tracking Benchmark}.
\newblock \emph{Int. J. Comput. Vis.}, 129:\penalty0 439--461, 2020.

\bibitem[Feng and Su(2024)]{Feng:2024}
Mingzheng Feng and Jianbo Su.
\newblock {RGBT Tracking: A Comprehensive Review}.
\newblock \emph{Inf. Fusion}, 110:\penalty0 102492, 2024.

\bibitem[Fu et~al.(2021)Fu, Liu, Fu, et~al.]{Fu:2021}
Zhihong Fu, Qingjie Liu, Zehua Fu, et~al.
\newblock {STMTrack: Template-free Visual Tracking with Space-time Memory
  Networks}.
\newblock \emph{IEEE/CVF Conf. Comput. Vis. Pattern Recog.}, pages
  13769--13778, 2021.

\bibitem[Gad et~al.(2022)Gad, Basmaji, Yaghi, et~al.]{Gad:2022}
Abdalla Gad, Tasnim Basmaji, Maha Yaghi, et~al.
\newblock {Multiple Object Tracking in Robotic Applications: Trends and
  Challenges}.
\newblock \emph{Applied Sciences}, 12\penalty0 (19), 2022.

\bibitem[Gunjal et~al.(2018)Gunjal, Gunjal, Shinde, et~al.]{Gunjal:2018}
Pramod~R. Gunjal, Bhagyashri~R. Gunjal, Haribhau~A. Shinde, et~al.
\newblock {Moving Object Tracking Using Kalman Filter}.
\newblock In \emph{Int. Conf. Adv. Commun. Comput. Technol.}, pages 544--547,
  2018.

\bibitem[Guo et~al.(2022)Guo, Wang, Yang, et~al.]{Guo:2022}
Shuman Guo, Shichang Wang, Zhenzhong Yang, et~al.
\newblock {A Review of Deep Learning-Based Visual Multi-Object Tracking
  Algorithms for Autonomous Driving}.
\newblock \emph{Applied Sciences}, 12\penalty0 (21), 2022.

\bibitem[He et~al.(2023)He, Li, Zhang, et~al.]{He:2023}
Chunming He, Kai Li, Yachao Zhang, et~al.
\newblock {Camouflaged Object Detection with Feature Decomposition and Edge
  Reconstruction}.
\newblock In \emph{IEEE/CVF Conference on Computer Vision and Pattern
  Recognition}, pages 22046--22055, 2023.

\bibitem[Hong et~al.(2024)Hong, Yan, Zhang, et~al.]{Hong:2024}
Lingyi Hong, Shilin Yan, Renrui Zhang, et~al.
\newblock {OneTracker: Unifying Visual Object Tracking with Foundation Models
  and Efficient Tuning}.
\newblock In \emph{IEEE/CVF Conf. Comput. Vis. Pattern Recog.}, pages
  19079--19091, 2024.

\bibitem[Huang et~al.(2021)Huang, Zhao, and Huang]{Huang:2021}
Lianghua Huang, Xin Zhao, and Kaiqi Huang.
\newblock {GOT-10k: A Large High-Diversity Benchmark for Generic Object
  Tracking in the Wild}.
\newblock \emph{IEEE Trans. Pattern Anal. Mach. Intell.}, 43\penalty0
  (5):\penalty0 1562--1577, 2021.

\bibitem[Jiang et~al.(2023)Jiang, Wang, Peng, et~al.]{Jiang:2023}
Nan Jiang, Kuiran Wang, Xiaoke Peng, et~al.
\newblock {Anti-UAV: A Large-Scale Benchmark for Vision-Based UAV Tracking}.
\newblock \emph{IEEE Trans. Multimedia}, 25:\penalty0 486--500, 2023.

\bibitem[Jin et~al.(2022)Jin, Lin, Wen, et~al.]{Jin:2022}
Ruibing Jin, Guosheng Lin, Changyun Wen, et~al.
\newblock {Feature Flow: In-network Feature Flow Estimation for Video Object
  Detection}.
\newblock \emph{Pattern Recognit.}, 122:\penalty0 108323, 2022.

\bibitem[Khan et~al.(2024)Khan, Dil, Alam, et~al.]{Khan:2024}
Misha~Urooj Khan, Mahnoor Dil, Muhammad~Zeshan Alam, et~al.
\newblock {SafeSpace MFNet: Precise and Efficient Multi-Feature Drone Detection
  Network}.
\newblock \emph{IEEE Trans. Veh. Technol.}, 73\penalty0 (3):\penalty0
  3106--3118, 2024.

\bibitem[Kim et~al.(2023)Kim, Kim, and Won]{Kim:2023}
Jun-Hwa Kim, Namho Kim, and Chee~Sun Won.
\newblock {High-Speed Drone Detection Based On Yolo-V8}.
\newblock In \emph{IEEE Int. Conf. Acoust., Speech, Signal Process.}, pages
  1--2, 2023.

\bibitem[Lenhard et~al.(2024)Lenhard, Weinmann, Jäger,
  et~al.]{Lenhard:2024_YOLOFEDER}
Tamara~R. Lenhard, Andreas Weinmann, Stefan Jäger, et~al.
\newblock {YOLO-FEDER FusionNet: A Novel Deep Learning Architecture for Drone
  Detection}.
\newblock In \emph{IEEE Int. Conf. Image Process.}, pages 2299--2305, 2024.

\bibitem[Lenhard et~al.(2025{\natexlab{a}})Lenhard, Weinmann, Franke,
  et~al.]{Lenhard:2024}
Tamara~R. Lenhard, Andreas Weinmann, Kai Franke, et~al.
\newblock {SynDroneVision: A Synthetic Dataset for Image-Based Drone
  Detection}.
\newblock In \emph{IEEE/CVF Winter Conf. Appl. Comput. Vis.}, pages 7637--7647,
  2025{\natexlab{a}}.

\bibitem[Lenhard et~al.(2025{\natexlab{b}})Lenhard, Weinmann, and
  Koch]{Lenhard:2025_YOLOFEDER}
Tamara~R. Lenhard, Andreas Weinmann, and Tobias Koch.
\newblock {Performance Optimization of YOLO-FEDER FusionNet for Robust Drone
  Detection in Visually Complex Environments}.
\newblock \emph{ArXiv}, abs/2509.14012, 2025{\natexlab{b}}.

\bibitem[Lenhard et~al.(2026)Lenhard, Weinmann, Snoussi, and
  Koch]{Lenhard:2025_RealDataDrones}
Tamara~R. Lenhard, Andreas Weinmann, Hichem Snoussi, and Tobias Koch.
\newblock {Long-Duration Drone Tracking Dataset}.
\newblock \emph{Zenodo}, 2026.
\newblock https://doi.org/10.5281/zenodo.17182190.

\bibitem[Li et~al.(2019)Li, Wu, Wang, et~al.]{Li:2018}
Bo Li, Wei Wu, Qiang Wang, et~al.
\newblock {SiamRPN++: Evolution of Siamese Visual Tracking With Very Deep
  Networks}.
\newblock In \emph{IEEE/CVF Conf. Comput. Vis. Pattern Recog.}, pages
  4277--4286, 2019.

\bibitem[Liu et~al.(2021)Liu, Fan, Ouyang, et~al.]{Liu:2021}
Hansen Liu, Kuangang Fan, Qinghua Ouyang, et~al.
\newblock {Real-Time Small Drones Detection Based on Pruned YOLOv4}.
\newblock \emph{Sensors}, 21\penalty0 (10), 2021.

\bibitem[Luo et~al.(2021)Luo, Xing, Milan, et~al.]{Luo:2021}
Wenhan Luo, Junliang Xing, Anton Milan, et~al.
\newblock {Multiple Object Tracking: A Literature Review}.
\newblock \emph{Artif. Intell.}, 293:\penalty0 103448, 2021.

\bibitem[Lv et~al.(2022)Lv, Ai, Chen, et~al.]{Lv:2022}
Yaowen Lv, Zhiqing Ai, Manfei Chen, et~al.
\newblock {High-Resolution Drone Detection Based on Background Difference and
  SAG-YOLOv5s}.
\newblock \emph{Sensors}, 22\penalty0 (15), 2022.

\bibitem[Marvasti-Zadeh et~al.(2022)Marvasti-Zadeh, Cheng, Ghanei-Yakhdan,
  et~al.]{Marvasti:2022}
Seyed~Mojtaba Marvasti-Zadeh, Li Cheng, Hossein Ghanei-Yakhdan, et~al.
\newblock {Deep Learning for Visual Tracking: A Comprehensive Survey}.
\newblock \emph{IEEE Trans. Intell. Transp. Syst.}, 23\penalty0 (5):\penalty0
  3943--3968, 2022.

\bibitem[Mohsan et~al.(2023)Mohsan, Othman, Li, et~al.]{Mohsan:2023}
Syed Agha~Hassnain Mohsan, Nawaf Qasem~Hamood Othman, Yanlong Li, et~al.
\newblock {Unmanned Aerial Vehicles (UAVs): Practical Aspects, Applications,
  Open Challenges, Security Issues, and Future Trends }.
\newblock \emph{Intell. Serv. Robot.}, 16:\penalty0 109--137, 2023.

\bibitem[Ravi et~al.(2024)Ravi, Gabeur, Hu, et~al.]{Ravi:2024}
Nikhila Ravi, Valentin Gabeur, Yuan-Ting Hu, et~al.
\newblock {SAM 2: Segment Anything in Images and Videos}.
\newblock \emph{ArXiv}, abs/2408.00714, 2024.

\bibitem[Redmon et~al.(2016)Redmon, Divvala, Girshick, et~al.]{Redmon:2015}
Joseph Redmon, Santosh~Kumar Divvala, Ross~B. Girshick, et~al.
\newblock {You Only Look Once: Unified, Real-Time Object Detection}.
\newblock \emph{IEEE/CVF Conf. Comput. Vis. Pattern Recog.}, pages 779--788,
  2016.

\bibitem[Singha and Aydin(2021)]{Singha:2021}
Subroto Singha and Burchan Aydin.
\newblock {Automated Drone Detection Using YOLOv4}.
\newblock \emph{Drones}, 5\penalty0 (3), 2021.

\bibitem[Svanstr{\"o}m et~al.(2021)Svanstr{\"o}m, Alonso-Fernandez, and
  Englund]{Svanstroem:2021}
Fredrik Svanstr{\"o}m, Fernando Alonso-Fernandez, and Cristofer Englund.
\newblock {A Dataset for Multi-Sensor Drone Detection}.
\newblock \emph{Data in Brief}, 39:\penalty0 107521, 2021.

\bibitem[Svanström et~al.(2022)Svanström, Alonso-Fernandez, and
  Englund]{Svanstroem:2022}
Fredrik Svanström, Fernando Alonso-Fernandez, and Cristofer Englund.
\newblock {Drone Detection and Tracking in Real-Time by Fusion of Different
  Sensing Modalities}.
\newblock \emph{Drones}, 6\penalty0 (11), 2022.

\bibitem[{Ultralytics}(accessed: 2026-01-05{\natexlab{a}})]{Ultralytics_YOLOv5}
{Ultralytics}.
\newblock {Ultralytics YOLOv5}.
\newblock \url{https://docs.ultralytics.com/models/yolov5/}, accessed:
  2026-01-05{\natexlab{a}}.

\bibitem[{Ultralytics}(accessed: 2026-01-05{\natexlab{b}})]{Ultralytics_YOLOv8}
{Ultralytics}.
\newblock {Explore Ultralytics YOLOv8}.
\newblock \url{https://docs.ultralytics.com/models/yolov8/}, accessed:
  2026-01-05{\natexlab{b}}.

\bibitem[Voigtlaender et~al.(2019)Voigtlaender, Luiten, Torr,
  et~al.]{Voigtlaender:2019}
Paul Voigtlaender, Jonathon Luiten, Philip H.~S. Torr, et~al.
\newblock {Siam R-CNN: Visual Tracking by Re-Detection}.
\newblock \emph{IEEE/CVF Conf. Comput. Vis. Pattern Recog.}, pages 6577--6587,
  2019.

\bibitem[Wang et~al.(2022{\natexlab{a}})Wang, Shi, Meng, et~al.]{Wang:2022}
Chuanyun Wang, Zhongrui Shi, Linlin Meng, et~al.
\newblock {Anti-Occlusion UAV Tracking Algorithm with a Low-Altitude Complex
  Background by Integrating Attention Mechanism}.
\newblock \emph{Drones}, 6\penalty0 (6), 2022{\natexlab{a}}.

\bibitem[Wang et~al.(2022{\natexlab{b}})Wang, Zhou, Xie, et~al.]{Wang:2021}
Xingjian Wang, Chengwei Zhou, Jiayang Xie, et~al.
\newblock {Drone Detection with Visual Transformer}.
\newblock In \emph{Int. Conf. on Auton. Unmanned Syst.}, pages 2689--2699,
  2022{\natexlab{b}}.

\bibitem[Yan et~al.(2019)Yan, Zhao, Wang, et~al.]{Bin:2019}
Bin Yan, Haojie Zhao, Dong Wang, et~al.
\newblock {‘Skimming-Perusal’ Tracking: A Framework for Real-Time and
  Robust Long-Term Tracking}.
\newblock In \emph{IEEE/CVF Int. Conf. Comput. Vis.}, pages 2385--2393, 2019.

\bibitem[Yang et~al.(2024)Yang, Huang, Chai, et~al.]{Yang:2024}
Cheng-Yen Yang, Hsiang-Wei Huang, Wenhao Chai, et~al.
\newblock {SAMURAI: Adapting Segment Anything Model for Zero-Shot Visual
  Tracking with Motion-Aware Memory}.
\newblock \emph{ArXiv}, abs/2411.11922, 2024.

\bibitem[Yao et~al.(2025)Yao, Guo, Yan, et~al.]{Yao:2025}
Siyuan Yao, Yang Guo, Yanyang Yan, et~al.
\newblock {UncTrack: Reliable Visual Object Tracking With Uncertainty-Aware
  Prototype Memory Network}.
\newblock \emph{IEEE Trans. Image Process.}, 34:\penalty0 3533--3546, 2025.

\bibitem[Yasmeen and Daescu(2025)]{Yasmeen:2025}
Arowa Yasmeen and Ovidiu Daescu.
\newblock {Recent Research Progress on Ground-to-Air Vision-Based Anti-UAV
  Detection and Tracking Methodologies: A Review}.
\newblock \emph{Drones}, 9\penalty0 (1), 2025.

\bibitem[Yun and Kim(2025)]{Yun:2025}
Sangyoon Yun and Hyunsoo Kim.
\newblock {YOLOv8-SAMURAI: A Hybrid Tracking Framework for Ladder Worker Safety
  Monitoring in Occlusion Scenarios}.
\newblock \emph{Buildings}, 15\penalty0 (11), 2025.

\bibitem[Zhang et~al.(2025)Zhang, Li, Liu, et~al.]{Zhang:2025}
Wenqi Zhang, Xinqiang Li, Xingyu Liu, et~al.
\newblock {Facing Challenges: A Survey of Object Tracking}.
\newblock \emph{Digit. Signal Process.}, 161:\penalty0 105082, 2025.

\bibitem[Zhao et~al.(2022)Zhao, Zhang, Li, et~al.]{Zhao:2022}
Jie Zhao, Jingshu Zhang, Dongdong Li, et~al.
\newblock {Vision-Based Anti-UAV Detection and Tracking}.
\newblock \emph{IEEE Trans. Intell. Transp. Syst.}, 23\penalty0 (12):\penalty0
  25323--25334, 2022.

\bibitem[Zheng et~al.(2020{\natexlab{a}})Zheng, Wang, Liu,
  et~al.]{Zheng:2020_1}
Zhaohui Zheng, Ping Wang, Wei Liu, et~al.
\newblock {Distance-IoU Loss: Faster and Better Learning for Bounding Box
  Regression}.
\newblock In \emph{AAAI Conf. Artif. Intell.}, pages 12993--13000,
  2020{\natexlab{a}}.

\bibitem[Zheng et~al.(2020{\natexlab{b}})Zheng, Wang, Ren,
  et~al.]{Zheng:2020_2}
Zhaohui Zheng, Ping Wang, Dongwei Ren, et~al.
\newblock {Enhancing Geometric Factors in Model Learning and Inference for
  Object Detection and Instance Segmentation}.
\newblock \emph{IEEE Trans. Cybern.}, 52:\penalty0 8574--8586,
  2020{\natexlab{b}}.

\bibitem[Zhou et~al.(2023)Zhou, Yang, Chen, et~al.]{Zhou:2023}
Xunkuai Zhou, Guidong Yang, Yizhou Chen, et~al.
\newblock {ADMNet: Anti-Drone Real-Time Detection and Monitoring}.
\newblock In \emph{IEEE/RSJ Int. Conf. Intell. Robots Syst.}, pages 3009--3016,
  2023.

\end{thebibliography}
}

\newpage
\twocolumn[{%
 \centering
 \vspace{1cm}
 \Large \textbf{Supplementary Material} \\[3.9em]
}]

\appendix
\setcounter{table}{0}
\setcounter{figure}{0}
\renewcommand{\thetable}{\Roman{table}}
\renewcommand{\thefigure}{\Roman{figure}}

\section{Sequence-Level Tracking Performance}
While the main paper primarily reports aggregated results on DUT Anti-UAV~\cite{Zhao:2022} and the custom datasets R1 and R2 (cf. Sec.~3.3, main paper), including comparisons with other tracking algorithms (cf. Tab.~\ref{tab:SAMURAI_zeroshot_on_DUT}), Tabs.~\ref{tab:SAMURAI_final_results_DUT} and \ref{tab:SAMURAI_final_results_custom} provide per-sequence evaluations. These results expose sequence-specific variations, offering a fine-grained characterization of SAMURAI and its detector-augmented extension.

\vspace{-0.3cm}
\paragraph{Performance on DUT Anti-UAV.} SAMURAI achieves stable tracking performance across most sequences, irrespective of whether initialization is based on ground-truth (GT) annotations or detector predictions (cf. GT vs. D, 2nd column, Tab.~\ref{tab:SAMURAI_final_results_DUT}). The detector-augmented extension further reinforces this stability, consistently matching or surpassing GT-based initialization and exhibiting robustness to initialization noise. In challenging sequences (\eg, \textit{video05}, \textit{video12}, \textit{video16}), detector-only initialization leads to noticeable performance degradation, whereas the detector-augmented variant mitigates these effects and recovers performance close to GT-level. 

\begin{figure*}
\centering        
  \includegraphics[width=0.87\textwidth, trim={0.3cm 0.8cm 0.3cm 0.2cm}, clip]{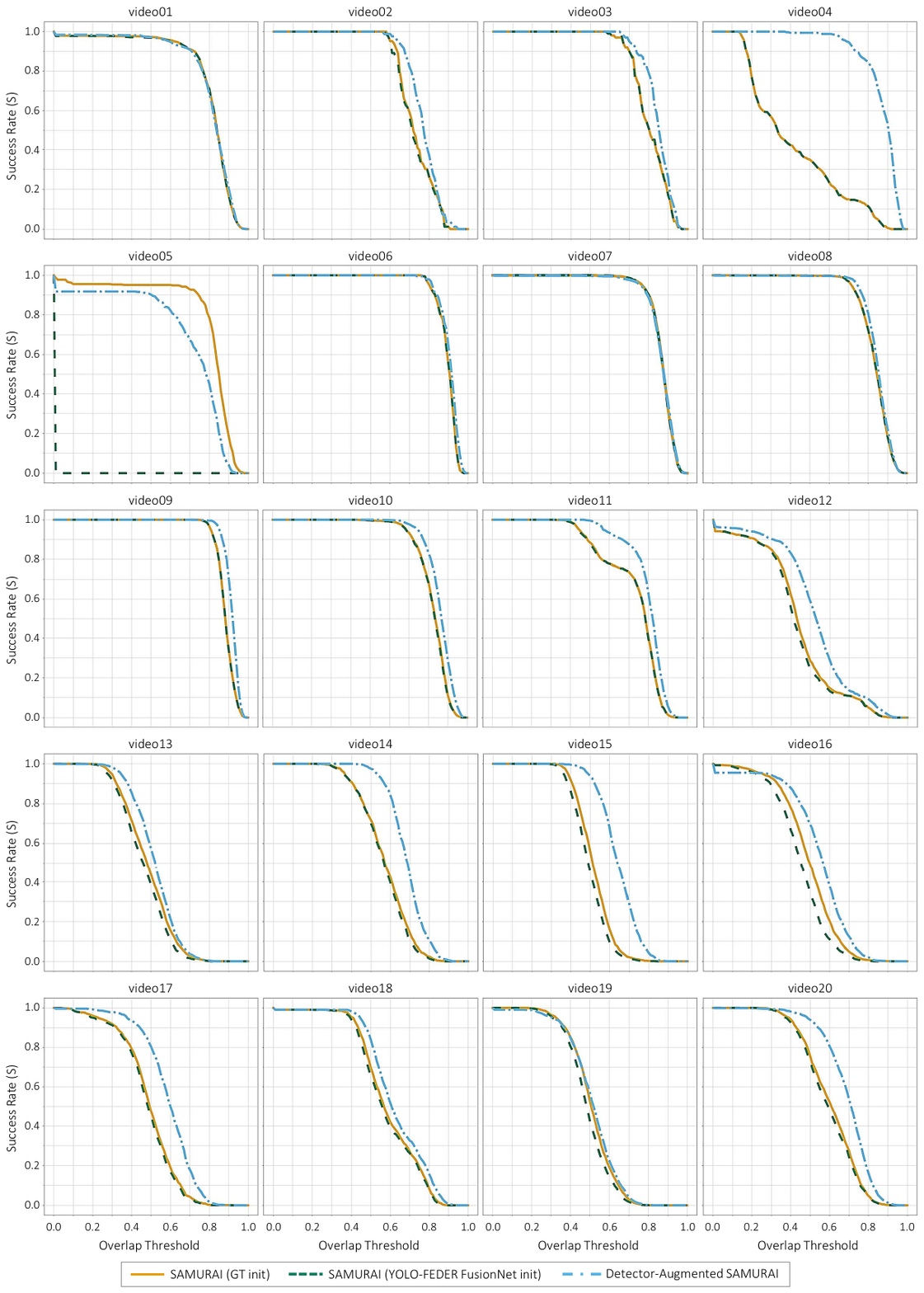}
  \caption{\label{fig:success_rate_DUTAntiUAV} Success plots on sequences from DUT Anti-UAV~\cite{Zhao:2022}, comparing SAMURAI with ground-truth initialization, SAMURAI with first-frame YOLO-FEDER FusionNet initialization, and the detector-augmented SAMURAI.}
\end{figure*}

\begin{figure*}
\centering
  \includegraphics[width=0.95\textwidth, trim={0cm 18.5cm 0cm 0cm}, clip]{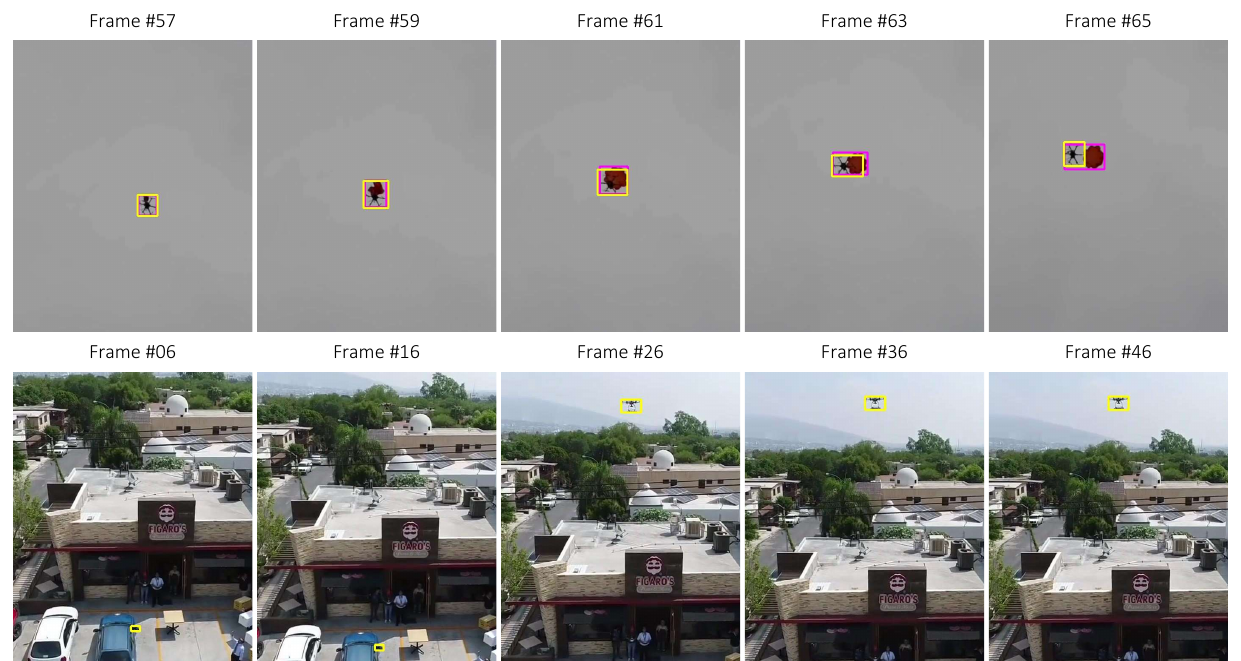}
  \captionof{figure}{\label{fig:example_1} Exemplary comparison between SAMURAI with first-frame YOLO-FEDER FusionNet initialization (magenta) and its detector-augmented extension (yellow), illustrating the benefits of continuous decoder-based prompting. Top row: First-frame initialization propagates erroneous masks under appearance variations (magenta), while continuous prompting corrects drift by re-aligning with detector outputs (yellow). Bottom row: First-frame initialization fails to recover from erroneous starting box, whereas detector-augmented SAMURAI leverages ongoing detections to reestablish accurate tracking (cf. frame 26, yellow).}
\end{figure*}

A representative example is \textit{video05}, where the initial detection erroneously marks the mirror of a car as the drone (cf. Fig.~\ref{fig:example_1}, bottom), while the actual drone is located at the top-center of the frame. In this situation, SAMURAI cannot recover from the erroneous initialization, whereas its detector-augmented extension leverages continuous detector feedback to correct the error and restore accurate tracking. Another example is \textit{video04}, where the unfolding of an attached parachute induces significant appearance changes, resulting in erroneous bounding boxes from inaccurate segmentation masks (cf. Fig.~\ref{fig:example_1}, top). With first-frame-only initialization (both GT- and detector-based), these errors accumulate and remain uncorrected. In contrast, detector augmentation enables SAMURAI to counteract these effects and sustain accurate tracking (cf. Fig.~\ref{fig:success_rate_DUTAntiUAV}).

On the other hand, sequences such as \textit{video06} and \textit{video10} (characterized by favorable tracking conditions, i.e., blue sky) achieve near-perfect performance across all metrics -- even under first-frame-only initialization via detector predictions (cf. Tab.~\ref{tab:SAMURAI_final_results_DUT}).

\begin{table}[t!]
\centering
\caption{\label{tab:SAMURAI_zeroshot_on_DUT} Comparison of SAMURAI (with GT initialization) against state-of-the-art trackers on the DUT Anti-UAV dataset~\cite{Zhao:2022}. Baseline results for all trackers (except SAMURAI) are reported as presented in~\cite{Zhao:2022}. Best results are in \textbf{bold}.}

\footnotesize
  \begin{tabular}{@{}|l|ccc|@{}}\hline
    Tracker      &S~$\uparrow$ & P~$\uparrow$ & $\text{P}_{\text{norm}}$~$\uparrow$\\\hline
    SiamFC~\cite{Bertinetto:2016} & 0.381 & 0.623 & 0.526\\\hline
    ECO~\cite{Danelljan:2017} & 0.404 & 0.717 & 0.643\\\hline
    SPLT~\cite{Bin:2019} & 0.405 & 0.651 & 0.585\\\hline
    SiamRPN++~\cite{Li:2018} & 0.545 & 0.780 & 0.709\\\hline
    ATOM~\cite{Danelljan:2019} & 0.578 & 0.830 & 0.758\\\hline
    DiMP~\cite{Bhat:2019} & 0.578 & 0.831 & 0.756\\\hline
    TransT~\cite{Chen:2021_TransT} & 0.586 & 0.832 & 0.765\\\hline
    LTMU~\cite{Dai:2020} & 0.608 & 0.783 & 0.858\\\hline
    \textbf{SAMURAI}~\cite{Yang:2024} & \textbf{0.663} & \textbf{0.888} & \textbf{0.973}\\\hline
  \end{tabular}
\end{table}

\begin{table*}
\centering\footnotesize
\caption{\label{tab:SAMURAI_final_results_DUT}Performance of SAMURAI and its detector-augmented extension (\ding{51}, 3rd column) on sequences from the DUT Anti-UAV dataset~\cite{Zhao:2022}. \textit{GT} (2nd column) denotes initialization with ground truth, while \textit{D} denotes detector-based initialization using the first YOLO-FEDER FusionNet prediction as the bounding-box prompt. Best results are highlighted in \textbf{bold}.}
  \begin{tabular}{@{}|l|c|c|ccc|ccc|c|c|@{}}
    \multicolumn{1}{c}{ } & \multicolumn{1}{c}{ } & \multicolumn{1}{c}{ }& \multicolumn{3}{c}{Tracking Metrics} & \multicolumn{5}{c}{Detection Metrics}\\
    \hline 
    Seq. & Init. & Detector & S~$\uparrow$ & P~$\uparrow$ & $\text{P}_{\text{norm}}$~$\uparrow$& \multicolumn{3}{c|}{mAP~$\uparrow$} & FNR~$\downarrow$ & FDR~$\downarrow$ \\
    & Method & Augmentation & & & & {\small @0.25} & {\small @0.5} & {\small @0.5-0.95} & & \\\hline
    \multirow{3}{*}{video01} & GT & -- & 0.808 & 0.878 & 0.959 & 0.989 & \textbf{0.983} & 0.697 & 0.021 & \textbf{0.000}\\
    &  D & -- & 0.807 & \textbf{0.878} & 0.959 & 0.989 & 0.981 & \textbf{0.697} & 0.022  & 0.001\\
    &  D & \ding{51} & \textbf{0.809} & 0.872 & \textbf{0.965} & \textbf{0.990} & 0.977 & 0.690 & \textbf{0.017} & 0.003\\\hline
    
    \multirow{3}{*}{video02} &  GT & -- & 0.730 & 0.926 & 0.980 & 0.995 & 0.995 & 0.466 & 0.000 & 0.000\\
    &  D & -- & 0.724 & 0.924 & 0.980 & 0.995 & 0.995 & 0.454 & 0.000 & 0.000\\
    &  D & \ding{51} & \textbf{0.769} & \textbf{0.941} & \textbf{0.980} & \textbf{0.995} & \textbf{0.995} &  \textbf{0.557} & \textbf{0.000} & \textbf{0.000} \\\hline

    \multirow{3}{*}{video03} &  GT & -- & 0.802 & 0.952 & 0.980 & 0.995 & 0.995 & 0.694 & 0.000 & 0.000\\
    &  D & -- & 0.802 & 0.951 & 0.980 & 0.995 & 0.995 & 0.694 & 0.000 & 0.000\\
    &  D & \ding{51} & \textbf{0.842} & \textbf{0.958} & \textbf{0.980} & \textbf{0.995} & \textbf{0.995} & \textbf{0.754} & \textbf{0.000} & \textbf{0.000}\\\hline
    
    \multirow{3}{*}{video04} &  GT & -- & 0.412 & 0.351 & 0.938 & 0.716 & 0.401 & 0.190 & 0.226 & 0.226\\
    &  D & -- & 0.412 & 0.352 & 0.938 & 0.716 & 0.401 & 0.190 & 0.226 & 0.226\\
    &  D & \ding{51} & \textbf{0.872} & \textbf{0.929} & \textbf{0.980} & \textbf{0.995} & \textbf{0.990} & \textbf{0.765} & \textbf{0.000} & \textbf{0.000}\\\hline
    
    \multirow{3}{*}{video05} &  GT & -- & \textbf{0.807} & \textbf{0.910} & \textbf{0.969}  & \textbf{0.939} & \textbf{0.933} & \textbf{0.679} & \textbf{0.044} & \textbf{0.040}\\
    &  D & -- & 0.010 & 0.000 & 0.014 & 0.000 & 0.000 & 0.000 & 1.000 & 1.000\\
    &  D & \ding{51} & 0.699 & 0.847 & 0.914 & 0.904 & 0.882 & 0.478 & 0.082 & 0.057\\\hline
    
    \multirow{3}{*}{video06} &  GT & -- & 0.893 & 0.935 & 0.980 & 0.995 & 0.995 & 0.817 & 0.000 & 0.000\\
    &  D & -- & 0.893 & 0.935 & 0.980 & 0.995 & 0.995 & 0.817 & 0.000 & 0.000\\
    &  D & \ding{51} & \textbf{0.904} & \textbf{0.947} & \textbf{0.980} & \textbf{0.995} & \textbf{0.995} & \textbf{0.841} & \textbf{0.000} & \textbf{0.000}\\\hline
    
    \multirow{3}{*}{video07} &  GT & -- & 0.868 & 0.905 & 0.980 & 0.943 & 0.943 & 0.728 & 0.000 & 0.073\\
    &  D & -- & \textbf{0.868} & 0.905 & \textbf{0.980} & \textbf{0.943} & \textbf{0.943} & 0.728 & \textbf{0.000} & \textbf{0.073}\\
    &  D & \ding{51} & 0.866 & \textbf{0.914} & 0.979 & 0.942 & 0.941 & \textbf{0.733} & 0.001 & 0.074\\\hline
    
    \multirow{3}{*}{video08} &  GT & -- & 0.834 & 0.933 & 0.980 & 0.969 & 0.967 & 0.683 & 0.000 & 0.028\\
    &  D & -- & 0.834 & 0.932 & \textbf{0.980} & 0.969 & 0.967 & 0.683 & 0.001 & 0.028\\
    &  D & \ding{51} & \textbf{0.846} & \textbf{0.936} & 0.979 & \textbf{0.969} & \textbf{0.967} & \textbf{0.708} & \textbf{0.001} & \textbf{0.028}\\\hline
    
    \multirow{3}{*}{video09} &  GT & -- & 0.879 & 0.938 & 0.980 & 0.898 & 0.898 & 0.713 & 0.000 & 0.109\\
    &  D & -- & 0.879 & 0.938 & 0.980 & 0.898 & 0.898 & 0.713 & 0.000 & 0.109\\
    &  D & \ding{51} & \textbf{0.912} & \textbf{0.946} & \textbf{0.980} & \textbf{0.898} & \textbf{0.898} & \textbf{0.776} & \textbf{0.000} & \textbf{0.109}\\\hline
    
    \multirow{3}{*}{video10} &  GT & -- & 0.820 & 0.917 & 0.980 & 0.995 & 0.993 & 0.682 & 0.000 & 0.002\\
    &  D & -- & 0.818 & 0.916 & 0.980 & 0.995 & 0.992 & 0.677 & 0.000 & 0.002\\
    &  D & \ding{51} & \textbf{0.857} & \textbf{0.929} & \textbf{0.980} & \textbf{0.995} & \textbf{0.995} & \textbf{0.753} & \textbf{0.000} & \textbf{0.002}\\\hline
    
    \multirow{3}{*}{video11} &  GT & -- & 0.729 &  0.917 & 0.980 & 0.994 & 0.846 & 0.464 & 0.000 & 0.002\\
    &  D & -- & 0.727 & 0.918 & 0.980 & 0.994 & 0.839 & 0.463 & 0.000 & 0.002\\
    &  D & \ding{51} & \textbf{0.797} & \textbf{0.938} & \textbf{0.980} & \textbf{0.994} & \textbf{0.990} & \textbf{0.611} & \textbf{0.000} & \textbf{0.002}\\\hline
    
    \multirow{3}{*}{video12} &  GT & -- & 0.438 &  0.850 & 0.932 & 0.905 & 0.203 & 0.065 & 0.093 & 0.065\\
    &  D & -- & 0.426 & 0.851 & 0.934 & 0.903 & 0.169 & 0.058 & 0.095 & 0.066\\
    &  D & \ding{51} & \textbf{0.523} & \textbf{0.895} & \textbf{0.958} & \textbf{0.940} & \textbf{0.524} & \textbf{0.139} & \textbf{0.060} & \textbf{0.047}\\\hline
    
    \multirow{3}{*}{video13} &  GT & -- & 0.479 & 0.922 & 0.980 & 0.987 & 0.374 & 0.076 & 0.001 & 0.001\\
    &  D & -- & 0.450 & 0.920 & \textbf{0.980} & 0.974 & 0.306 & 0.058 & 0.002 & 0.003\\
    &  D & \ding{51} & \textbf{0.517} & \textbf{0.936} & 0.979 & \textbf{0.993} & \textbf{0.510} & \textbf{0.105} & \textbf{0.001} & \textbf{0.001} \\\hline
    
    \multirow{3}{*}{video14} &  GT & -- & 0.568 & 0.906 & 0.980 & 0.994 & 0.731 & 0.218 & 0.000 & 0.000\\
    &  D & -- & 0.560 & 0.904 & 0.980 & 0.995 & 0.705 & 0.196 & 0.000 & 0.000\\
    &  D & \ding{51} & \textbf{0.682} & \textbf{0.922} & \textbf{0.980} & \textbf{0.995} & \textbf{0.975} & \textbf{0.396} & \textbf{0.000} & \textbf{0.000}\\\hline
    
    \multirow{3}{*}{video15} &  GT & -- & 0.516 & 0.933 & 0.980 & 0.988 & 0.519 & 0.095 & 0.000 & 0.014\\
    &  D & -- & 0.494 & 0.930 & 0.980 & 0.988 & 0.381 & 0.066 & 0.000 & \textbf{0.014}\\
    &  D & \ding{51} & \textbf{0.642} & \textbf{0.939} & \textbf{0.980} & \textbf{0.988} & \textbf{0.925} & \textbf{0.310} & \textbf{0.000} & 0.015\\\hline
    
    \multirow{3}{*}{video16} &  GT & -- & 0.491 & 0.903 & 0.974 & 0.942 & 0.374 & 0.083 & \textbf{0.032} & 0.026\\
    &  D & -- & 0.449 & \textbf{0.903} & \textbf{0.974} & 0.922 & 0.235 & 0.046 & 0.041 & 0.034\\
    &  D & \ding{51} & \textbf{0.542} & 0.889 & 0.937 & \textbf{0.975} & \textbf{0.628} & \textbf{0.162} & 0.045 & \textbf{0.001}\\\hline
    
    \multirow{3}{*}{video17} &  GT & -- & 0.488 & 0.908 & 0.979 & 0.941 & 0.440 & 0.095 & 0.040 & 0.039\\
    &  D & -- & 0.478 & 0.907 & \textbf{0.979} & 0.934 & 0.405 & 0.085 & 0.050 & 0.045\\
    &  D & \ding{51} & \textbf{0.591} & \textbf{0.923} & 0.977 & \textbf{0.983} & \textbf{0.791} & \textbf{0.229} & \textbf{0.008} & \textbf{0.004}\\\hline
    
    \multirow{3}{*}{video18} &  GT & -- & 0.590 & 0.938 & 0.971 & 0.995 & 0.632 & 0.203 & 0.009 & 0.000\\
    &  D & -- & 0.580 & 0.937 & 0.971 & 0.994 & 0.580 & 0.188 & 0.009 & 0.000\\
    &  D & \ding{51} & \textbf{0.628} & \textbf{0.950} & \textbf{0.971} & \textbf{0.995} & \textbf{0.806} & \textbf{0.268} & \textbf{0.009} & \textbf{0.000}\\\hline
    
    \multirow{3}{*}{video19} &  GT & -- & 0.508 & 0.934 & 0.980 & \textbf{0.993} & 0.485 & 0.099 & \textbf{0.002} & 0.002\\
    &  D & -- & 0.487 & 0.934 & \textbf{0.980} & 0.990 & 0.380 & 0.073 & 0.003 & 0.003\\
    &  D & \ding{51} & \textbf{0.512} & \textbf{0.935} & 0.971 & 0.984 & \textbf{0.583} & \textbf{0.129} & 0.011 & \textbf{0.002}\\\hline
    
    \multirow{3}{*}{video20} &  GT & -- & 0.601 & 0.910 & 0.980 & 0.995 & 0.692 & 0.218 & 0.000 & 0.000\\
    &  D & -- & 0.590 & 0.907 & \textbf{0.980} & \textbf{0.995} & 0.660 & 0.199 & \textbf{0.000} & \textbf{0.000}\\
    &  D & \ding{51} & \textbf{0.691} & \textbf{0.941} & 0.979 & 0.994 & \textbf{0.935} & \textbf{0.389} & 0.002 & 0.001\\\hline
  \end{tabular}
\end{table*}

\begin{figure*}
\centering
  \includegraphics[width=0.95\textwidth, trim={0cm 18.5cm 0cm 0cm}, clip]{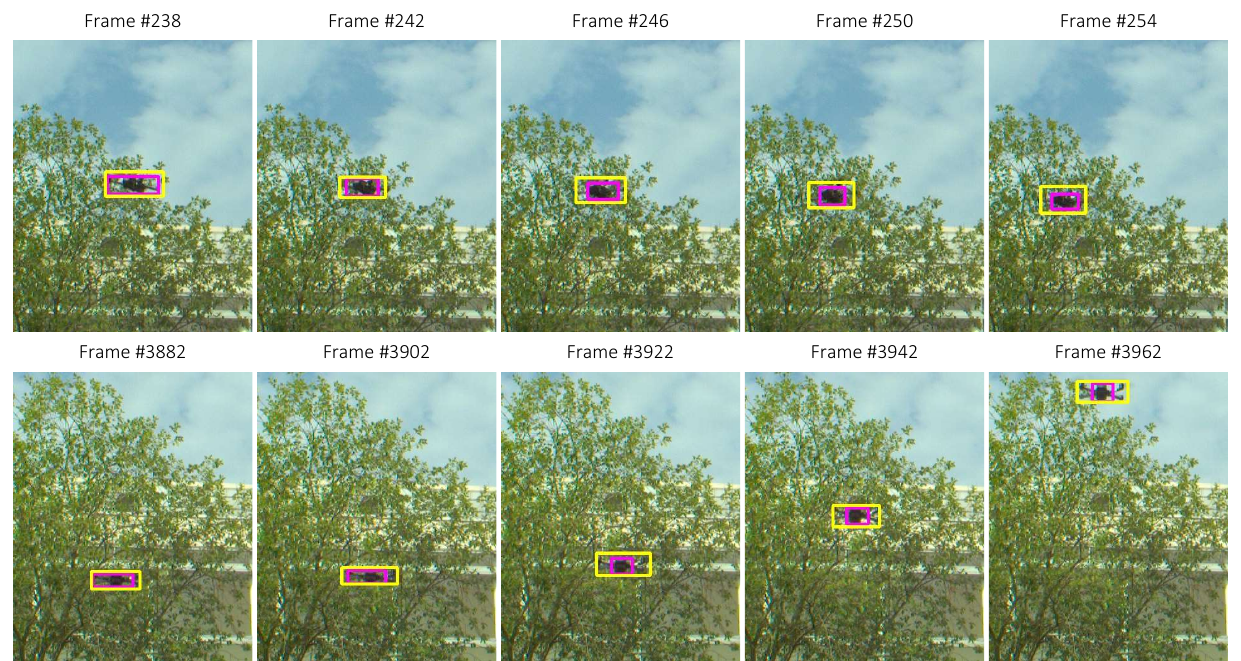}
  \captionof{figure}{\label{fig:example_2} Comparison between SAMURAI with first-frame-only initialization (magenta) and detector-augmented SAMURAI (yellow) with bounding-box averaging. The detector-augmented variant remains accurate in textured regions (\eg, tree crowns), whereas first-frame initialization tends to degrade to partial drone-body predictions. }
\end{figure*}

\begin{figure*}
\centering
  \includegraphics[width=0.95\textwidth, trim={0cm 24cm 0cm 0cm}, clip]{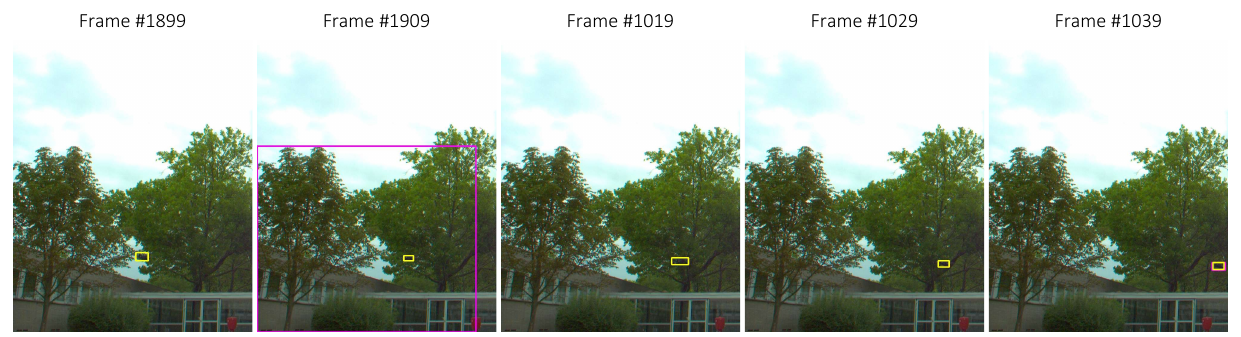}
  \captionof{figure}{\label{fig:example_3} SAMURAI with first-frame YOLO-FEDER FusionNet initialization (magenta) yields unstable (occasionally oversized) predictions when the drone leaves and reenters the FOV, whereas the detector-augmented SAMURAI (yellow) maintains robust tracking.}
\end{figure*}

\vspace{-0.3cm}
\paragraph{Performance on Custom Data.}
On the custom datasets R1 and R2, SAMURAI exhibits pronounced sequence-level performance variations. In the \textit{POS3} sequences of both datasets, detector-only initialization without detector-based augmentation leads to substantial degradation, with low success rates, reduced mAP values, and elevated FNRs (cf. Tab.~\ref{tab:SAMURAI_final_results_custom} and Fig.~\ref{fig:example_3}). However, when leveraging the detector-augmented version of SAMURAI, performance improves markedly: tracking scores often double, and detection quality rises to levels comparable to or even exceeding GT initialization. For instance, in \textit{POS3} (R1), the success rate increases from 0.289 to 0.560, while the FNR is reduced by more than half. Visual inspection (cf. Fig.~\ref{fig:example_2}) further reveals that the observed improvements in mAP are driven not only by continuous prompting through detector-derived bounding boxes but also by the averaging mechanism embedded in the proposed Prediction Fusion Module (cf. Sec.~3.2).

\begin{table*}
\centering
\caption{\label{tab:SAMURAI_final_results_custom}Performance of SAMURAI and its detector-augmented extension (\ding{51}, 4th column) on sequences from R1 and R2. \textit{GT} (3rd column) denotes initialization with ground truth, while \textit{D} denotes detector-based initialization using the first YOLO-FEDER FusionNet prediction as the bounding-box prompt. Best results are highlighted in \textbf{bold}.}

\footnotesize
  \begin{tabular}{@{}|l|l|c|c|ccc|ccc|c|c|@{}}
    \multicolumn{1}{c}{ } & \multicolumn{1}{c}{ } & \multicolumn{1}{c}{ }& \multicolumn{1}{c}{ }& \multicolumn{3}{c}{Tracking Metrics} & \multicolumn{5}{c}{Detection Metrics}\\
    \hline
    Dataset & Seq. & Init. & Detector & S~$\uparrow$ & P~$\uparrow$ & $\text{P}_{\text{norm}}$~$\uparrow$& \multicolumn{3}{c|}{mAP~$\uparrow$} & FNR~$\downarrow$ & FDR~$\downarrow$ \\
    & & Method & Augmentation & & & & {\small @0.25} & {\small @0.5} & {\small @0.5-0.95} & & \\\hline
    \multirow{6}{*}{R1} & \multirow{3}{*}{POS3} & GT & -- & 0.124 & 0.200 & 0.213 & 0.584 & 0.390 & 0.120 & 0.784 & 0.007\\
    && D & -- & 0.289 & 0.508 & 0.548 & 0.740 & 0.395 & 0.115 & 0.448 & 0.013 \\
    && D & \ding{51} & \textbf{0.560} & \textbf{0.762} & \textbf{0.808} & \textbf{0.907} & \textbf{0.792} & \textbf{0.352} & \textbf{0.179} & \textbf{0.005}\\\cline{2-12}
    
    & \multirow{3}{*}{POS7} & GT & -- & 0.360 & 0.551 & 0.591 & 0.793 & 0.597 & 0.187 & 0.398 & 0.002\\
    && D & -- & 0.397 & 0.622 & 0.668 & 0.829 & 0.596 & 0.177 & 0.319 & 0.002\\
    && D & \ding{51} & \textbf{0.710} & \textbf{0.874} & \textbf{0.930} & \textbf{0.970} & \textbf{0.914} & \textbf{0.499} & \textbf{0.052} & \textbf{0.002}\\\hline
    
     \multirow{6}{*}{R2} & \multirow{3}{*}{POS3} & GT & -- & 0.178 & 0.339 & 0.670 & 0.480 & 0.071 & 0.012 & 0.614 & 0.631 \\
    && D & -- & 0.168 & 0.344 & 0.658 & 0.481 & 0.044 & 0.007 & 0.621 & 0.636\\
    && D & \ding{51} & \textbf{0.454} & \textbf{0.833} & \textbf{0.888} & \textbf{0.934} & \textbf{0.427} & \textbf{0.101} & \textbf{0.116} & \textbf{0.026} \\\cline{2-12}
    
    & \multirow{3}{*}{POS7} & GT & -- & 0.578 & 0.846 & 0.972 & 0.994 & 0.726 & 0.236 & 0.009 & 0.001\\
    && D & -- & 0.580 & \textbf{0.846} & \textbf{0.972} & \textbf{0.994} & 0.730 & 0.239 & \textbf{0.009} & \textbf{0.001}\\
    && D & \ding{51} & \textbf{0.684} & 0.831 & 0.966 & 0.984 & \textbf{0.861} & \textbf{0.427} & 0.015 & 0.007\\\hline
  \end{tabular}
\end{table*}

\section{Qualitative Analysis of Detector-Augmented SAMURAI Limitations}
\begin{figure}
\centering
  \includegraphics[width=0.48\textwidth, trim={0.7cm 18.2cm 9.5cm 0.1cm}, clip]{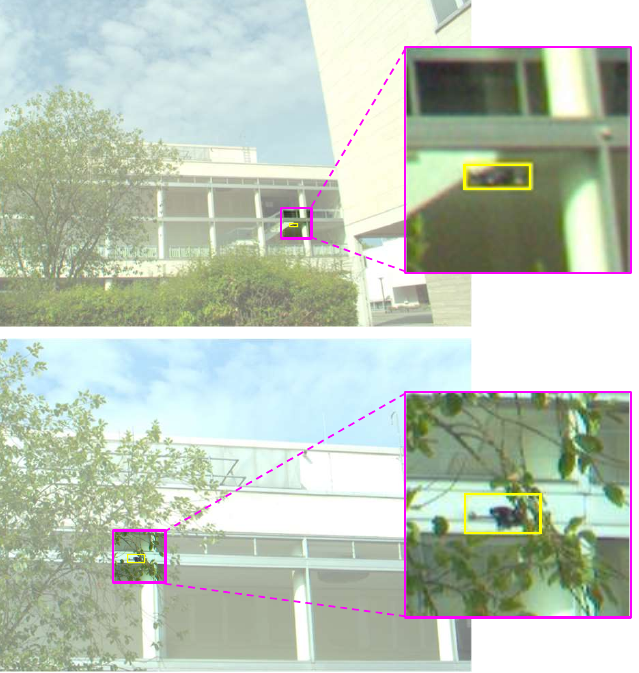}
  \captionof{figure}{\label{fig:TrackingFailureExamples} Qualitative examples of representative tracking failure cases for detector-augmented SAMURAI. Zoomed-in regions are shown on the right-hand side to enhance the visibility of small-scale objects. Yellow bounding boxes denote GT drone localization. (top: R1; bottom: R2)}
\end{figure}

Fig.~\ref{fig:TrackingFailureExamples} presents representative tracking failure cases of SAMURAI's detector-augmented extension on R1 (top) and R2 (bottom). These failures predominantly occur in scenarios where drone targets appear at very small scales or are partially occluded, resulting in limited visual evidence for reliable association and thus in tracking instabilities.

\section{Detection Performance}
\label{sec:results_YOLOFEDER}
Tabs.~\ref{tab:YOLO-FEDER_final_results_1} and~\ref{tab:YOLO-FEDER_final_results_2} report the sequence-level detection performance of YOLO-FEDER FusionNet~\cite{Lenhard:2024_YOLOFEDER,Lenhard:2025_YOLOFEDER} on the custom datasets R1 and R2, as well as on the publicly available DUT Anti-UAV dataset~\cite{Zhao:2022} (tracking subset). On R1 and R2, the detector achieves consistently high mAP with low FNRs and FDRs for \textit{POS7}, whereas \textit{POS3} exhibits comparatively higher FNRs. Results on the DUT Anti-UAV dataset confirm this trend: most sequences achieve near-perfect detection at lower IoU thresholds and retain competitive performance under stricter evaluation. Except for \textit{video12} and \textit{video16}, both FNR and FDR remain consistently low across sequences. Overall, YOLO-FEDER FusionNet provides promising detection performance across diverse conditions.

\begin{table}[t!]
\caption{YOLO-FEDER FusionNet performance on R1 and R2.}
\label{tab:YOLO-FEDER_final_results_1}
\footnotesize
  \centering
  \begin{tabular}{@{}|l|l|ccc|c|c|@{}}
    \hline
    Data- & Seq. & \multicolumn{3}{c|}{mAP~$\uparrow$} & FNR~$\downarrow$ & FDR~$\downarrow$ \\
    set& & {\small @0.25} & {\small @0.5} & {\small @0.5-0.95} & & \\\hline
    \multirow{2}{*}{R1} & POS3 & 0.782 & 0.759 & 0.342 & 0.312 & 0.058\\\cline{2-7}
    & POS7 & 0.944 & 0.925 &  0.503 & 0.088 & 0.046\\\hline
    \multirow{2}{*}{R2} & POS3 & 0.902 & 0.433 & 0.117 & 0.214 & 0.007\\\cline{2-7}
    & POS7 & 0.983 & 0.906 & 0.449 & 0.078 & 0.013\\\hline
  \end{tabular}
\end{table}

When compared to SAMURAI with GT initialization, YOLO-FEDER FusionNet achieves superior performance, particularly on the custom datasets R1 and R2 (cf. Tabs.~\ref{tab:SAMURAI_final_results_custom} and \ref{tab:YOLO-FEDER_final_results_1}). However, in combination with SAMURAI -- also referred to as detector-augmented SAMURAI -- additional improvements are obtained beyond standalone YOLO-FEDER FusionNet. While gains in bounding-box localization are modest, with mAP values comparable to or slightly exceeding those of YOLO-FEDER FusionNet, the most significant benefit is reflected in FNR, with reductions of up to 41.99\% on R1 and R2.

\begin{table}[t!]
\centering\footnotesize
\caption{YOLO-FEDER FusionNet performance on DUT Anti-UAV~\cite{Zhao:2022}.}
\label{tab:YOLO-FEDER_final_results_2}
  \begin{tabular}{@{}|l|ccc|c|c|@{}}
    \hline
   Seq. & \multicolumn{3}{c|}{mAP~$\uparrow$} & FNR~$\downarrow$ & FDR~$\downarrow$ \\
    & {\small @0.25} & {\small @0.5} & {\small @0.5-0.95} & & \\\hline

    video01 & 0.972 & 0.964 & 0.661 & 0.059 & 0.012\\\hline
    video02 & 0.995 & 0.995 & 0.569 & 0.000 & 0.000\\\hline
    video03 & 0.995 & 0.995 & 0.743 & 0.000 & 0.000\\\hline
    video04 & 0.995 & 0.995 & 0.829 & 0.006 & 0.015\\\hline
    video05 & 0.967 & 0.942 & 0.500 & 0.056 & 0.025\\\hline
    video06 & 0.995 & 0.995 & 0.861 & 0.000 & 0.000\\\hline
    video07 & 0.994 & 0.994 & 0.783 & 0.008 & 0.006\\\hline
    video08 & 0.995 & 0.995 & 0.729 & 0.003 & 0.000\\\hline
    video09 & 0.995 & 0.995 & 0.844 & 0.000 & 0.010\\\hline
    video10 & 0.995 & 0.995 & 0.740 & 0.000 & 0.020\\\hline
    video11 & 0.993 & 0.993 & 0.684 & 0.000 & 0.000\\\hline
    video12 & 0.933 & 0.601 & 0.184 & 0.113 & 0.065 \\\hline
    video13 & 0.982 & 0.549 & 0.109 & 0.060 & 0.012\\\hline
    video14 & 0.995 & 0.986 & 0.390 & 0.000 & 0.031\\\hline
    video15 & 0.995 & 0.964 & 0.330 & 0.002 & 0.001\\\hline
    video16 & 0.919 & 0.761 & 0.199 & 0.157 & 0.018\\\hline
    video17 & 0.983 & 0.827 & 0.228 & 0.060 & 0.034 \\\hline
    video18 & 0.995 & 0.837 & 0.335 & 0.010 & 0.005\\\hline
    video19 & 0.995 & 0.605 & 0.134 & 0.000 & 0.003\\\hline
    video20 & 0.995 & 0.964 & 0.441 & 0.006 & 0.042\\\hline
  \end{tabular}
\end{table}

\begin{figure}
\centering
  \includegraphics[width=0.48\textwidth, trim={0cm 15cm 10.5cm 0cm}, clip]{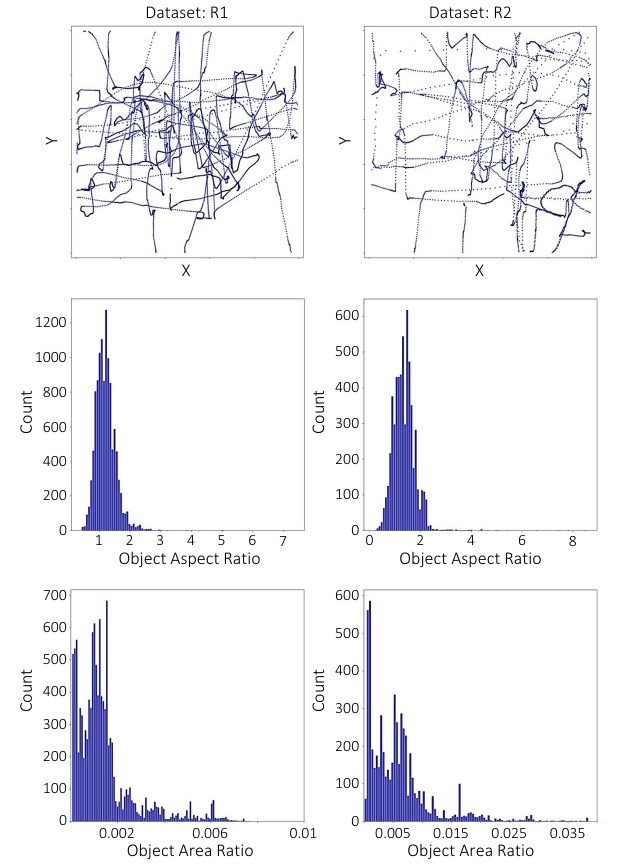}
  \captionof{figure}{\label{fig:R1_R2_Characteristics} Distributions of drone positions (top), object aspect ratios (middle), and object area ratios (bottom) across all sequences in R1 and R2.}
\end{figure}

\begin{figure*}
\centering
  \includegraphics[width=0.94\textwidth, trim={0.5cm 20cm 0.5cm 0cm}, clip]{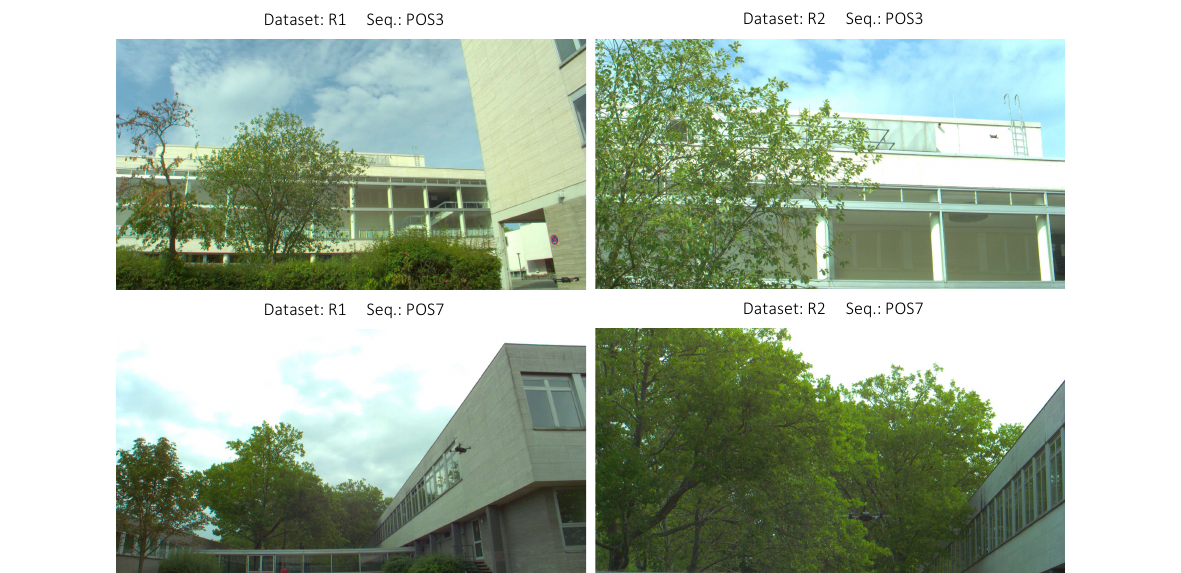}
  \captionof{figure}{\label{fig:R1_R2_FOV} Representative frames from the four custom-recorded sequences in R1 and R2, highlighting the sequence-level FOV and the visual attributes of the surrounding environment.}
\end{figure*}

\begin{figure*}
\centering
  \includegraphics[width=0.94\textwidth, trim={0.7cm 15cm 0.7cm 0cm}, clip]{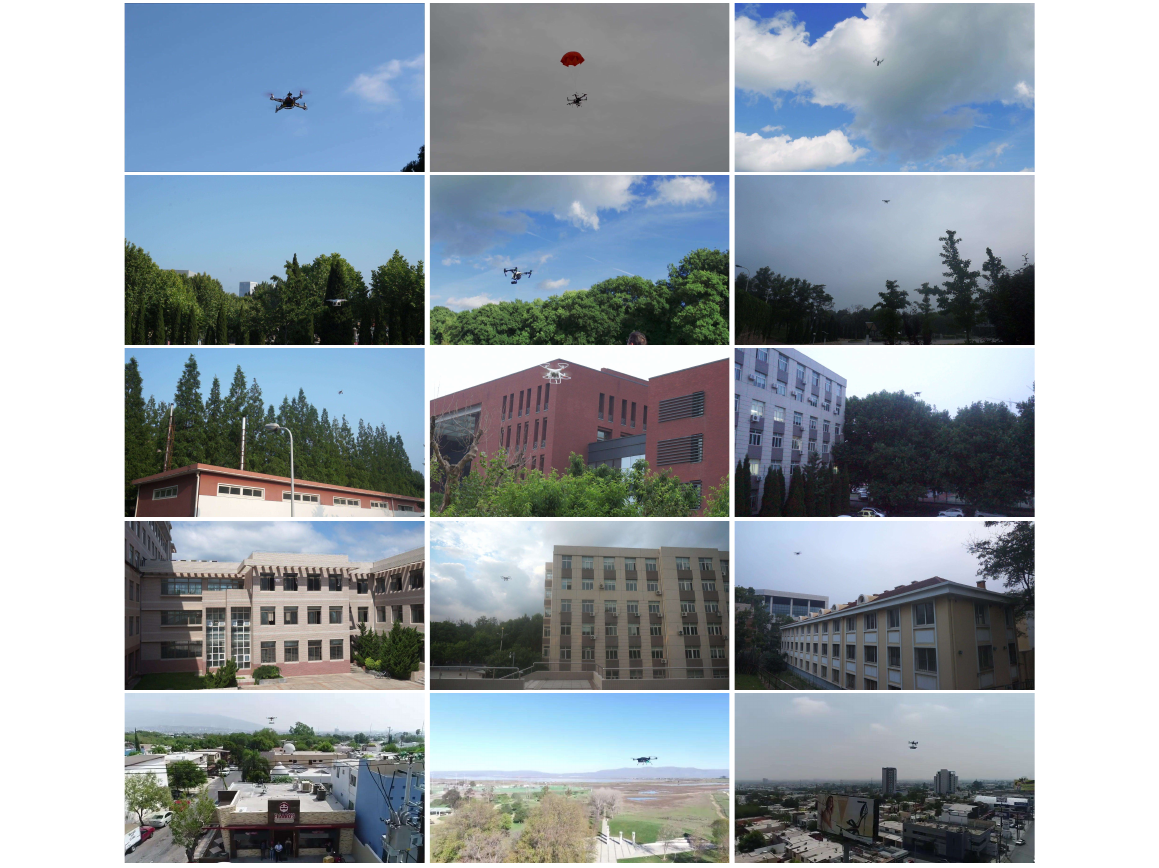}
  \captionof{figure}{\label{fig:DUT_Variation} Visual diversity of the DUT Anti-UAV dataset~\cite{Zhao:2022}.}
\end{figure*}

\section{Custom Dataset Details}
The custom-recorded datasets R1 and R2 (cf. Sec.~3.3, main paper) encompass urban outdoor environments with varying structural compositions, combining architectural features, vegetation, and open-sky regions (see Fig.~\ref{fig:R1_R2_FOV}). R1 is characterized by more pronounced architectural structure, such as multi-story facades, accompanied by moderate vegetation. In contrast, R2 contains denser foliage and greater natural clutter, with buildings more frequently occluded, resulting in visually richer and more complex scenes that are predominantly vegetation-driven. This makes R2 especially valuable for drone-detection research, as identifying drones against highly textured vegetative backgrounds is inherently challenging due to reduced visual contrast and pronounced camouflage effects~\cite{Lenhard:2024_YOLOFEDER}. Beyond the environmental variations, the drones in R1 and R2 also exhibit distinct spatial and geometric properties (see Fig.~\ref{fig:R1_R2_Characteristics}): R1 features denser and more complex motion patterns, with drone instances tending toward more compact aspect ratios and slightly larger relative areas. Conversely, R2 features a broader aspect-ratio distribution and smaller area ratios. Both datasets are released as part of this work and are publicly available at \cite{Lenhard:2025_RealDataDrones}.

\section{Visual Variability of DUT Anti-UAV}
The publicly available DUT Anti-UAV dataset encompasses diverse outdoor environments, ranging from sky-dominant scenes (Fig.~\ref{fig:DUT_Variation}, 1st row) to forested areas with dense vegetation (2nd row), as well as suburban and urban settings featuring varied architectural elements (rows 3-4). It captures multiple drone types from diverse viewpoints -- varying in angle, distance, and altitude -- leading to substantial changes in scale and appearance. Overall, DUT Anti-UAV exhibits significant diversity in drone appearance, background texture, illumination, and overall scene complexity.

\end{document}